\documentclass[runningheads]{llncs_cr}
\usepackage{graphicx}

\usepackage{amsmath,amssymb}
\usepackage{color}

\usepackage{subcaption}
\usepackage{xcolor}
\usepackage{url}
\usepackage{float}

\begin{document}

\title{A Temporally-Aware Interpolation Network for Video Frame Inpainting} 
\titlerunning{A Temporally-Aware Interpolation Network for Video Frame Inpainting} 


\author{Ximeng Sun\thanks{indicates equal contribution} \and
Ryan Szeto\textsuperscript{$\star$} \and
Jason J. Corso
}
%

\authorrunning{X. Sun et al.} 


\institute{University of Michigan, Ann Arbor, USA \\
\email{\{sunxm, szetor, jjcorso\}@umich.edu}}

\maketitle

\begin{abstract}

We propose the first deep learning solution to video frame inpainting, a more challenging but less ambiguous task than related problems such as general video inpainting, frame interpolation, and video prediction. We devise a pipeline composed of two modules: a bidirectional video prediction module and a temporally-aware frame interpolation module. The prediction module makes two intermediate predictions of the missing frames, each conditioned on the preceding and following frames respectively, using a shared convolutional LSTM-based encoder-decoder. The interpolation module blends the intermediate predictions, using time information and hidden activations from the video prediction module to resolve disagreements between the predictions. Our experiments demonstrate that our approach produces more accurate and qualitatively satisfying results than a state-of-the-art video prediction method and many strong frame inpainting baselines. Our code is available at \url{https://github.com/sunxm2357/TAI_video_frame_inpainting}.

\end{abstract}

\section{Introduction}
\label{sec:introduction}

\begin{figure}[t]
    \centering
    \includegraphics[width=\textwidth]{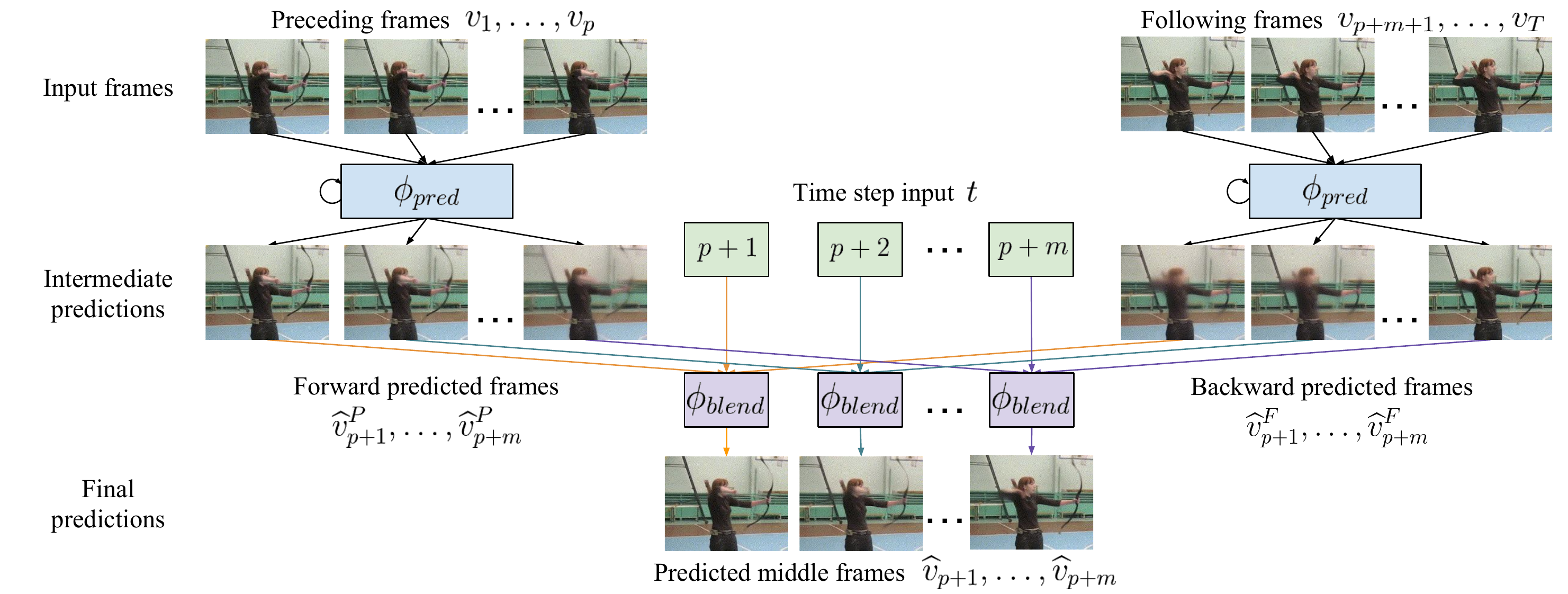}
    \caption{
    We predict middle frames by blending forward and backward intermediate video predictions with a Temporally-Aware Interpolation (TAI) network}
    \label{fig:model_arch}
\end{figure}

In this work, we explore the video frame inpainting problem, i.e. the task of reconstructing a missing sequence of frames given both a sequence of \textit{preceding} frames and a sequence of \textit{following} frames. For example, given a clip of someone winding up a baseball pitch and a clip of that person after he/she has released the ball, we would predict the clip of that person throwing the ball.
%
This task is more challenging than general video inpainting because we aim to fill in whole, temporally-contiguous frames rather than small spatio-temporal regions. It is also less ambiguous---and therefore more well-defined and easier to evaluate---than frame interpolation and video prediction, where methods cannot access the contextual information required to rule out many plausible predictions.

We present the first deep neural network for video frame inpainting, which approaches the problem in two steps as shown in Fig.~\ref{fig:model_arch}. First, we use a video prediction subnetwork to generate two intermediate predictions of the middle frames: the ``forward prediction'' conditioned on the preceding frames, and the ``backward prediction'' conditioned on the following frames. Then, we blend each pair of corresponding frames together to obtain the final prediction.

Our blending strategy exploits three characteristics of our intermediate prediction process. First, a pair of intermediate prediction frames for the same time step might be inconsistent, e.g. an actor might appear in two different locations. To address this, we introduce a blending neural network that can cleanly merge a given pair of predictions by reconciling the differences between them. Second, for any given time step, the forward and backward predictions are not equally reliable: the former is more accurate for earlier time steps, and the latter is more accurate for later time steps. Hence, we feed time step information directly into the blending network, making it \textit{temporally-aware} by allowing it to blend differently depending on which time step it is operating at. Finally, the intermediate predictions come from a neural network whose hidden features may be useful for blending. To leverage these features, we feed them to the blending network as additional inputs. We call our blending module the \textbf{Temporally-Aware Interpolation Network} (or TAI network for short). As we show in our experiments, our approach yields the most visually satisfying predictions among several strong baselines and across multiple human action video datasets.

In summary, we make the following contributions. First, we propose a deep neural network for video frame inpainting that generates two intermediate predictions and blends them with our novel TAI network. Second, we propose and compare against several baselines that leverage the information provided by the preceding and following frames, but do not utilize our TAI blending approach. Finally, we demonstrate that our approach is quantitatively and qualitatively superior to the proposed baselines across three human action video datasets.

\section{Related Work}
\label{sec:related_work}

In the \textit{general video inpainting problem} (of which our video frame inpainting task is a challenging instance), we are given a video that is missing arbitrary voxels (spatio-temporal pixels), and the goal is to fill each voxel with the correct value. Existing methods generally fall into one of three categories: \textit{patch-based} methods that search for complete spatio-temporal patches to copy into the missing area~\cite{wexler2004space,jia2005video,shen2006video,newson2014video}; \textit{object-based} methods that separate spatial content into layers (e.g. foreground and background), repair them individually, and stitch them back together~\cite{patwardhan2007video,jia2004video}; and \textit{probabilistic model-based} methods that assign values that maximize the likelihood under some probabilistic model~\cite{cheung2008video,granados2012background,ebdelli2015video}. Many of these approaches make strong assumptions about the video content, such as constrained camera pose/motion~\cite{shen2006video,patwardhan2007video,jia2004video} or static backgrounds~\cite{patwardhan2007video,jia2004video,granados2012background}. In addition, they are designed for the case in which ``holes'' in the video are localized to small spatio-temporal regions, and may therefore perform poorly when whole, contiguous frames are missing. Finally, to the best of our knowledge, no existing solution has leveraged deep neural networks, which can potentially outperform prior work thanks to the vast amounts of video data available online.

In the \textit{frame interpolation task}, the goal is to predict one or more frames in between two (typically subsequent) input frames. While most classical approaches linearly interpolate a dense optical flow field to an arbitrary number of intermediate time steps~\cite{borzi2003optimal,chen2011image,werlberger2011optical}, recent approaches train deep neural networks to predict one intermediate frame~\cite{long2016learning,niklaus2017bvideo,liu2017video}. However, all of these approaches require input frames that occur within a miniscule window of time (i.e. no more than 0.05 seconds apart), whereas we are interested in predicting on larger time scales. Furthermore, the task is ambiguous because a pair of individual frames without temporal context cannot sufficiently constrain the appearance of the intermediate frames (for instance, if we observed two frames of a swinging pendulum, we would need its period of oscillation to rule out several plausible appearances). As a result, it is hard to evaluate plausible predictions that deviate from the actual data.

\textit{Video prediction}, where the goal is to generate the future frames that follow a given sequence of video frames, is yet another actively-studied area with an important limitation. The earliest approaches draw heavily from language modeling literature by extending simple recurrent sequence-to-sequence models to predict patches of video~\cite{ranzato2014video,srivastava2015unsupervised}; more recent methods utilize structured models that decompose the input data and/or the learned representations in order to facilitate training~\cite{lotter2016deep,kalchbrenner2016video,villegas2017decomposing}. As with frame interpolation, video prediction is inherently underconstrained since the past can diverge into multiple plausible futures.

\section{Approach} \label{sec:approach}

\subsection{Problem Statement} \label{sub_sec:prob_state}
We define the video frame inpainting problem as follows. Let $V = \left \{v_1, v_2, \dots, v_T \right \}$ be a sequence of frames from a real video, $p$, $m$, and $f$ be the number of ``preceding'', ``middle'', and ``following'' frames such that $p+m+f=T$, and $P_V=\left \{v_1,\dots,v_p \right \}, M_V=\left \{v_{p+1},\dots,v_{p+m} \right \}, F_V= \left \{v_{p+m+1},\dots,v_{T} \right \}$ be the sequences of preceding, middle, and following frames from $V$ respectively. We seek a function $\phi$ that satisfies $M_V = \phi \left( P_V,F_V \right)$ for all $V$.

\subsection{Model Overview}
\label{sec:model_overview}

We propose a novel deep neural network to approximate the video inpainting function $\phi$ (see Fig.~\ref{fig:model_arch}). Instead of learning a direct mapping from the preceding and following sequences to the middle sequence, our model decomposes the problem into two sub-problems and tackles each one sequentially with two tractable modules: the Bidirectional Video Prediction Network (Sec.~\ref{sub_sec:video_pred}) and the Temporally-Aware Interpolation Network (Sec.~\ref{sub_sec:blend_module}).

\begin{itemize}
\item The \textbf{Bidirectional Video Prediction Network} generates two intermediate predictions of the middle sequence $M_V$, where each prediction is conditioned solely on the preceding sequence $P_V$ and the following sequence $F_V$ respectively.

\item The \textbf{Temporally-Aware Interpolation Network} blends corresponding frames from the predictions made by the Bidirectional Video Prediction Network, thereby producing the final prediction $\widehat{M}_V$. It accomplishes this by leveraging intermediate activations from the Bidirectional Video Prediction Network, as well as scaled time steps that explicitly indicate the relative temporal location of each frame in the final prediction.
\end{itemize}

\noindent Even though our model factorizes the video frame inpainting process into two steps, it is optimized in an end-to-end manner.

\subsection{Bidirectional Video Prediction Network $\phi_{pred}$} \label{sub_sec:video_pred}

\begin{figure}[t]
	\begin{subfigure}[b]{0.5\textwidth}
		\centering
        \includegraphics[width=\textwidth]{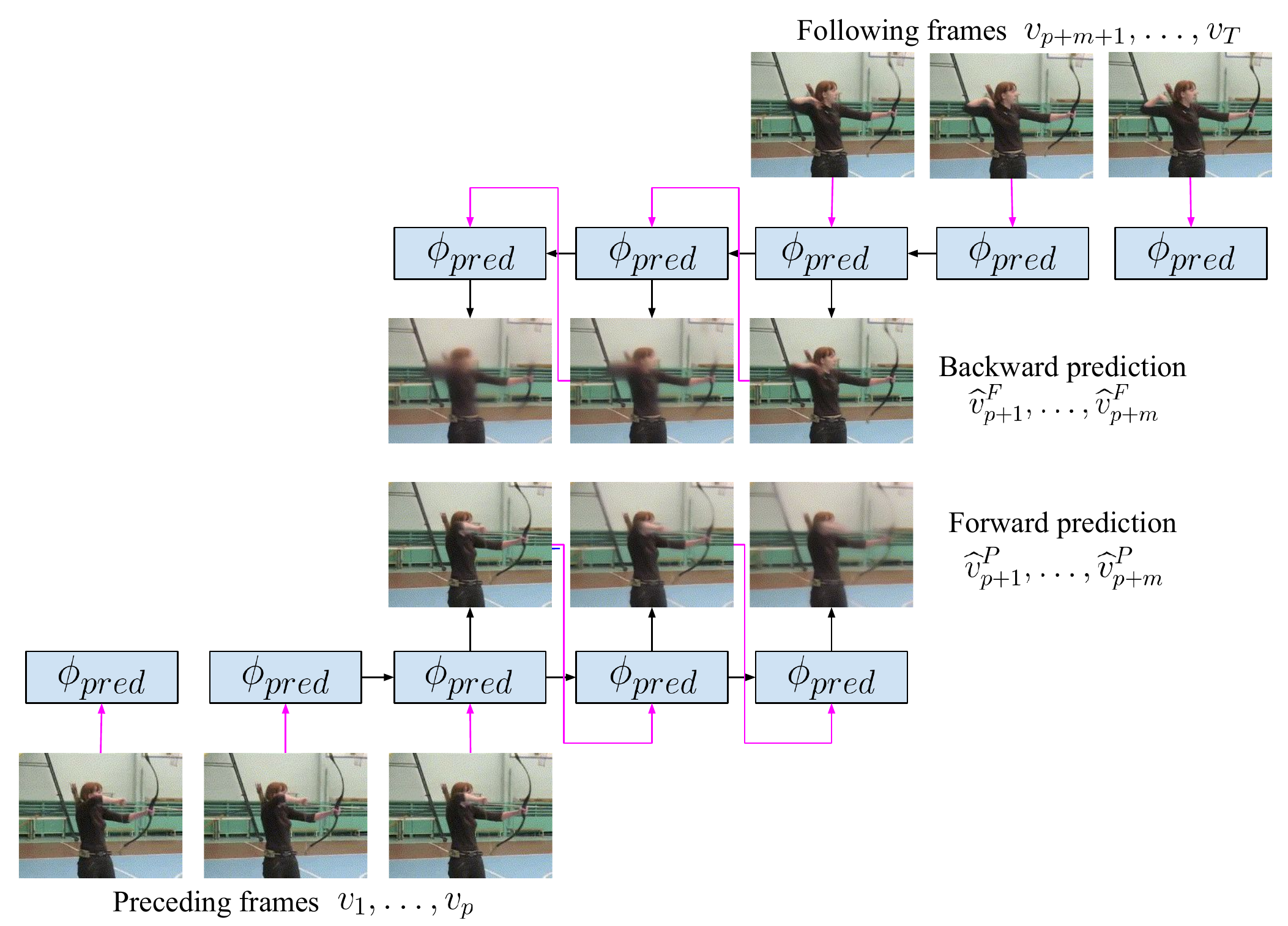}
        \caption{The Bidirectional Video Prediction Network}
        \label{fig:mcnet}
	\end{subfigure}
	\begin{subfigure}[b]{0.5\textwidth}
		\centering
        \includegraphics[width=\textwidth]{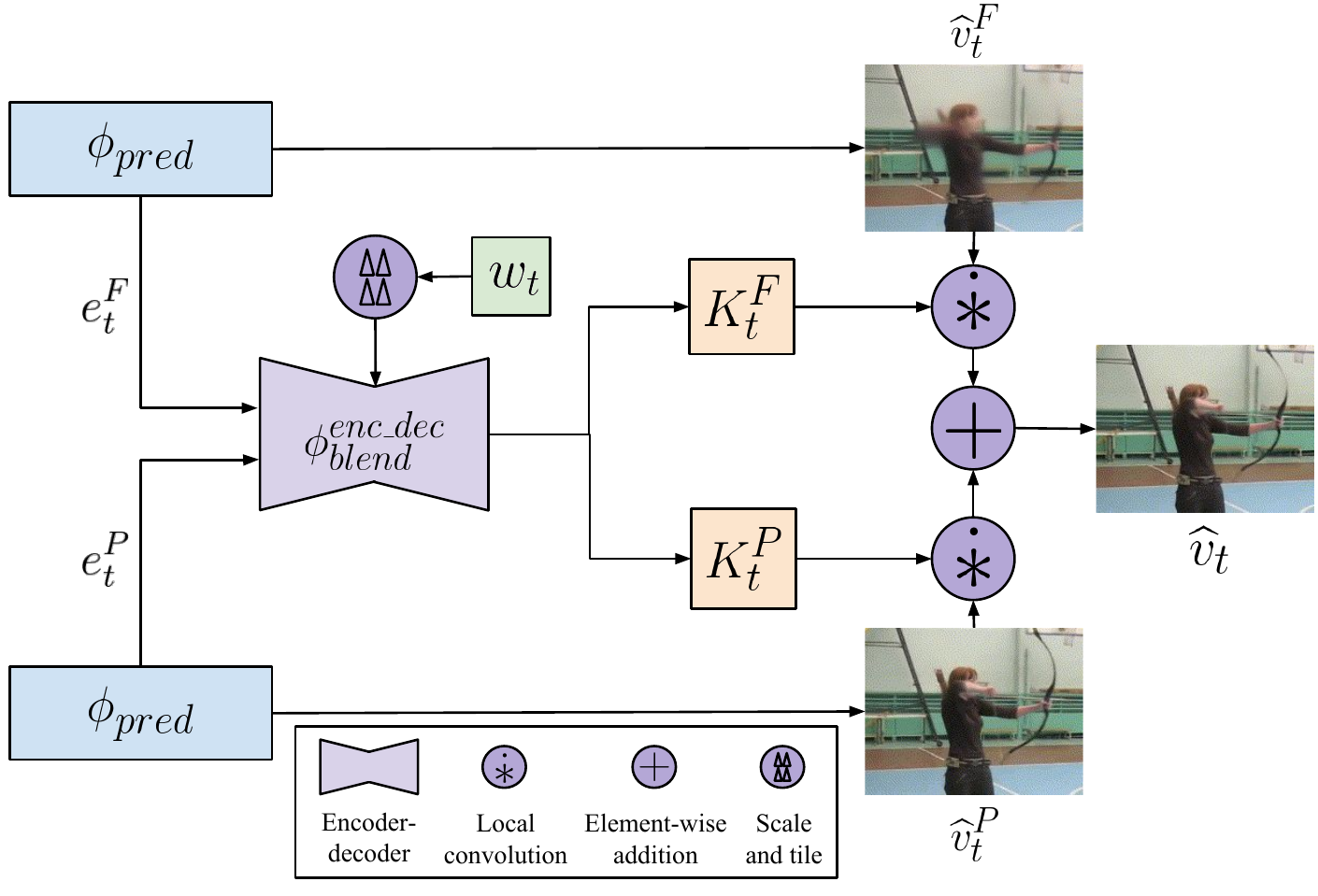}
        \caption{TAI network}
        \label{fig:TAI_arch}
	\end{subfigure}
    \caption{Architectures of two modules in our model}
\end{figure}

We first use the Bidirectional Video Prediction Network $\phi_{pred}$, shown in Fig.~\ref{fig:mcnet}, to produce two intermediate predictions---a ``forward prediction'' $\widehat{M}_V^P = \{ \widehat{v}_{p+1}^P, \dots, \\ \widehat{v}_{p+m}^P \}$ and a ``backward prediction'' $\widehat{M}_V^F = \{ \widehat{v}_{p+1}^F, \dots, \widehat{v}_{p+m}^F \}$---by conditioning on the preceding sequence $P_V$ and the following sequence $F_V$ respectively:
\begin{align}
\widehat{M}_V^P &= \phi_{pred} \left( P_V \right) \, , \\
\widehat{M}_V^F &= \left[\phi_{pred} \big((F_V)^R \big) \right]^R \, ,
\end{align}
where $(\cdot)^R$ is an operation that reverses the input sequence. Note that the same parameters are used to generate the forward and backward predictions.

In particular, the Bidirectional Video Prediction Network recurrently generates one frame at a time by conditioning on all previously generated frames. For example, in the case of the forward prediction:
\begin{align}
    \widehat{v}_{k+1}^P = \phi_{pred} \big( \{\tilde{v}_{1}^P, \tilde{v}_{2}^P, \dots, \tilde{v}_{k}^P\} \big) \, ,
\end{align}
where for a given $t$, $\tilde{v}_t^P$ is either $v_t$ (an input frame) if $t \in \{1, \dots, p\}$ or $\widehat{v}_t^P$ (an intermediate predicted frame) if $t \in \{p+1, \dots, p+m\}$. We also store intermediate activations from the Bidirectional Video Prediction Network (denoted as $e_t^P$), which serve as inputs to the Temporally-Aware Interpolation Network. We apply an analogous procedure to obtain the backward prediction and its corresponding intermediate activations.

\subsection{Temporally-Aware Interpolation Network $\phi_{blend}$} \label{sub_sec:blend_module}

Following the Bidirectional Video Prediction Network, the Temporally-Aware Interpolation Network $\phi_{blend}$ takes corresponding pairs of frames from $\widehat{M}_V^P$ and $\widehat{M}_V^F$ with the same time step, i.e. $\left( \widehat{v}_t^P, \widehat{v}_t^F \right)$ for each time step $t \in \{ p+1, \dots, p+m \}$, and blends them into the frames that make up the final prediction $\widehat{M}_V$:
\begin{align}
\widehat{v}_t &= \phi_{blend} \left( \widehat{v}_t^P, \widehat{v}_t^F \right) \, , \\
\widehat{M}_V &= \{\widehat{v}_t \mid t = p+1, \dots, p+m \} \, .
\end{align}

Blending $\widehat{v}_t^P$ and $\widehat{v}_t^F$ is difficult because (i) they often contain mismatched content (e.g. between the pair of frames, objects might be in different locations), and (ii) they are not equally reliable (e.g. $\widehat{v}_t^P$ is more reliable for earlier time steps). As we show in Sec.~\ref{sec:results}, equally averaging $\widehat{v}_t^P$ and $\widehat{v}_t^F$ predictably results in ghosting artifacts (e.g. multiple faded limbs in human action videos), but remarkably, replacing a simple average with a state-of-the-art interpolation network also exhibits the same problem.

In order to blend corresponding frames more accurately, our Temporally-Aware Interpolation (TAI) Network utilizes two additional sources of information. Aside from the pair of frames to blend, it receives the scaled time step to predict---defined as $w_t = (t-p)/(m+1)$---and the intermediate activations from the Bidirectional Video Prediction Network $e_t^P$ and $e_t^F$. We feed $w_t$ to our interpolation network so it can learn how to incorporate the unequal reliability of $\widehat{v}_t^P$ and $\widehat{v}_t^F$ into its final prediction; we feed $e_t^P$ and $e_t^F$ to leverage the high-level semantics that the Bidirectional Video Prediction Network has learned. We contrast standard interpolation with TAI algebraically:
\begin{align}
    \widehat{v}_t &= \phi_{interp} \left( \widehat{v}_t^P, \widehat{v}_t^F \right) \, , \label{eq:interp_eq} \\
    \widehat{v}_t &= \phi_{TAI} \left( \widehat{v}_t^P, e_t^P, \widehat{v}_t^F, e_t^F, w_t \right) \, . \label{eq:tai_eq}
\end{align}

TAI blends pairs of intermediate frames $(\widehat{v}_t^P, \widehat{v}_t^F)$ by first applying a unique, adaptive 2D kernel to each patch in the two input frames, and then summing the resulting images pixel-wise \textit{\`a la} Niklaus et al.~\cite{niklaus2017bvideo}. To generate the set of adaptive kernels, we use an encoder-decoder model, shown in Fig.~\ref{fig:TAI_arch}, that takes in the intermediate activations from the Bidirectional Video Prediction Network, $e_t^P$ and $e_t^F$, and the scaled time step $w_t$:
\begin{align}
    K^P_{t}, K^F_{t} = \phi^{enc\_dec}_{blend} \left( e_t^P, e_t^F, w_t \right) \, ,
\end{align}
where $K^P_{t}$ and $K^F_{t}$ are 3D tensors whose height and width match the frame resolution and whose depth equals the number of parameters in each adaptive kernel. Note that we inject the scaled time step by replicating it spatially and concatenating it to the output of one of the decoder's hidden layers as an additional channel. Afterwards, we apply the adaptive kernels to each input frame and sum the resulting images pixel-wise:
\begin{align}
    \widehat{v}_t(x,y) = K^P_{t}(x,y) * \mathcal{P}_P(x,y) + K^F_{t}(x,y) * \mathcal{P}_F(x,y) \, ,
\end{align}
where $\widehat{v}_t(x,y)$ is the pixel value of the final prediction at $(x, y)$, $K_t^{(\cdot)}(x,y)$ is the 2D kernel parameterized by the depth column of $K_t^{(\cdot)}$ at $(x, y)$, $*$ is the convolution operator, and $\mathcal{P}_{(\cdot)}(x,y)$ is the patch centered at $(x, y)$ in $\widehat{v}_t^{(\cdot)}$.

\subsection{Training Strategy}

To train our complete video frame inpainting model, we use both reconstruction-based and adversarial objective functions, the latter of which has been shown by Mathieu et al.~\cite{mathieu2015deep} to improve the sharpness of predictions. In our case, we train a discriminator $D$, which is a binary classification CNN, to distinguish between clips from the dataset and clips generated by our model. Meanwhile, we train our model---the ``generator''---to not only fool the discriminator, but also generate predictions that resemble the ground truth.

We update the generator and the discriminator in an alternating fashion. In the generator update step, we update our model by minimizing the following structured loss:
\begin{align}
	\mathcal{L}_{g} &= \alpha \bigg[ \mathcal{L}_{img}\left(\widehat{M}_V^P, M_V\right) + \mathcal{L}_{img}\left(\widehat{M}_V^F, M_V\right) + \mathcal{L}_{img}\left(\widehat{M}_V, M_V\right) \bigg] \nonumber \\
    &\quad {} + \beta \mathcal{L}_{GAN}\left(\widehat{M}_V\right) \, , \label{eq:generator_loss} \\
	\mathcal{L}_{GAN}\left(\widehat{M}_V\right) &= - \log D\left(\big[P_V, \widehat{M}_V, F_V\big]\right),
\end{align}
where $\alpha$ and $\beta$ are hyperparameters to balance the contribution of the reconstruc\-tion-based loss $\mathcal{L}_{img}$ and the adversarial loss $\mathcal{L}_{GAN}$. Note that we supervise the final prediction $\widehat{M}_V$ as well as the intermediate predictions $\widehat{M}_V^P$ and $\widehat{M}_V^F$ simultaneously. The loss $\mathcal{L}_{img}$ consists of the squared-error loss $\mathcal{L}_2$ and the image gradient difference loss $\mathcal{L}_{gdl}$~\cite{mathieu2015deep}, which encourages sharper predictions by penalizing differences along the edges in the image:
\begin{align}
	\mathcal{L}_{img}\left(\widehat{M}_V^{(\cdot)}, M_V\right) &= \mathcal{L}_2\left(\widehat{M}_V^{(\cdot)}, M_V\right)+ \mathcal{L}_{gdl}\left(\widehat{M}_V^{(\cdot)}, M_V\right) \, , \\
    \label{eq:L2}
    \mathcal{L}_2\left(\widehat{M}_V^{(\cdot)}, M_V\right) &= \sum\limits_{t=p+1}^{p+m} \left\| v_t - \widehat{v}_t^{(\cdot)} \right\|_2^{2} \, , \\
    \label{eq:Lgdl}
    \mathcal{L}_{gdl}\left(\widehat{M}_V^{(\cdot)}, M_V\right) &= \sum\limits_{t=p+1}^{p+m} \sum\limits_{i,j}^{h,w} \bigg( \Big| \big| v_t(i,j) - v_t(i-1,j)  \big| - \big| \widehat v_t^{(\cdot)}(i,j) - \widehat v_t^{(\cdot)}(i-1,j) \big| \Big| \nonumber   \\
    &\quad {} + \Big| \big| v_t(i,j-1) - v_t(i,j) \big| - \big| \widehat v_t^{(\cdot)}(i,j-1) - \widehat v_t^{(\cdot)}(i,j) \big| \Big| \bigg) \, .
\end{align}
Here, $\widehat{M}_V^{(\cdot)}$ can be one of the intermediate predictions $\left\{ \widehat{M}_V^P, \widehat{M}_V^F \right \}$ or the final prediction $\widehat{M}_V$.
In the discriminator update step, we update the discriminator by minimizing the cross-entropy error:
\begin{align}
	\mathcal{L}_{d} = - \log D \Big (\big[P_V, M_V, F_V \big] \Big) - \log \bigg(1- D\left(\big[P_V, \widehat{M}_V, F_V\big]\right)\bigg) \, .
\end{align}
We use the same discriminator as Villegas et al.~\cite{villegas2017decomposing}, but replace each layer that is followed by batch normalization~\cite{ioffe15batch} with a spectral normalization layer~\cite{miyato2018spectral}, which we have found results in more accurate predictions.

\section{Experiments}
\label{sec:results}

\subsection{Experimental Setup}

Our high-level approach to video frame inpainting places few constraints on the network architectures that can be used to implement each module (Sec.~\ref{sec:model_overview}). We instantiate the Bidirectional Video Prediction Network with MC-Net~\cite{villegas2017decomposing}. As for the Temporally-Aware Interpolation Network, we modify the Separable Adaptive Kernel Network~\cite{niklaus2017bvideo} to take as input intermediate activations and scaled time steps (refer to the supplementary materials for architectural details). These choices afford us two benefits: (i) the chosen networks are, to the best of our knowledge, the best-performing models in their original tasks, enabling us to demonstrate the full potential of our approach; and (ii) both networks are fully-convolutional, allowing us to modify the video resolution at test time.

We compare our video frame inpainting model to several baselines (Sec.~\ref{sec:baselines}) on videos from three human action datasets: KTH Actions~\cite{schuldt2004recognizing}, HMDB-51~\cite{kuehne11hmdb}, and UCF-101~\cite{soomro2012ucf101}. KTH Actions contains a total of 2,391 grayscale videos with resolution 120~$\times$~160 (height~$\times$~width) across six action classes; it also provides a standard training and testing set. We divide the standard training set into a smaller training set and a validation set, which are used for training and hyperparameter search respectively. Following Villegas et al.~\cite{villegas2017decomposing}, we reduce the resolution to 128~$\times$~128. We train each model to predict five middle frames from five preceding and five following frames; at inference time, we evaluate each model on its ability to predict ten middle frames from five preceding and five following frames. We double the number of frames to predict at test time in order to evaluate generalization performance (following Villegas et al.~\cite{villegas2017decomposing}).

HMDB-51 contains 6,849 RGB videos across 51 action classes; each video has a fixed height of 240 pixels. The dataset provides three cross-validation folds (each including a training and a test set); we take the test videos from the first fold as our test set and separate the remaining videos into our training and validation sets. During training, we reduce the resolution of each video to 160~$\times$~208, and train each model to predict three middle frames from four preceding and four following frames. At test time, we scale all videos to 240~$\times$~320 resolution (following Villegas et al.~\cite{villegas2017decomposing}) and take in the same number of preceding/following frames, but predict five frames in the middle.

UCF-101 contains 13,320 RGB videos with resolution 240~$\times$~320 across 101 action classes. It provides three cross-validation folds for action recognition; we take the test videos from the first fold as our test set and divide the remaining videos into our training and validation sets. The remainder of our experimental setup for UCF-101 matches our setup for HMDB-51.

\subsection{Baselines}
\label{sec:baselines}

The first baseline we compare our method to is MC-Net~\cite{villegas2017decomposing}---we re-implement and train their model to predict the middle frames conditioned only on the preceding frames. We also introduce two classes of baselines specifically designed for the video frame inpainting problem. In the first class, instead of learning a function $\phi$, we hand-craft several $\phi$'s that can perform well on certain video prediction tasks, particularly on videos with little movement or periodic motion. The baselines described by Eqs.~\ref{trivial_baseline_1}-\ref{trivial_baseline_3} copy or take a simple average of the last preceding frame $v_p$ and the first following frame $v_{p+m+1}$:
\begin{align}
	\phi_{\mathrm{repeat\_P}}\left(P_V,F_V\right) &= \left\{v_p, v_p, \dots, v_p\right\} \, , \label{trivial_baseline_1} \\
    \phi_{\mathrm{repeat\_F}}\left(P_V,F_V\right) &= \left\{v_{p+m+1}, v_{p+m+1}, \dots, v_{p+m+1}\right\} \, , \label{trivial_baseline_2} \\
    \phi_{\mathrm{SA\_P\_F}}\left(P_V,F_V\right) &= \left\{\widehat{v}, \widehat{v}, \dots, \widehat{v} \right\}, \mathrm{ where } \ \widehat{v} = \frac{v_p + v_{p+m+1}}{2}  \label{trivial_baseline_3} \, .
\end{align}
Also, we try incorporating the scaled time step of the predicted frame $w_t=(t-p)/(m+1)$ by computing a time-weighted average of $v_p$ and $v_{p+m+1}$:
\begin{align}
    \phi_{\mathrm{TW\_P\_F}} \left( P_V,F_V \right) &= \left\{\widehat{v}_{p+1}, \widehat{v}_{p+2}, \dots, \widehat{v}_{p+m}\right\}  \, , \label{trivial_baseline_4} \\
    \widehat{v}_t &= \left( 1-w_t \right) v_p + w_t v_{p+m+1} \, .
\end{align}
In the second class of baselines, we highlight the value of our TAI module by proposing two bidirectional prediction models that use the same Bidirectional Video Prediction Network architecture as our full model, but blend the forward and backward predictions without an interpolation network. Instead, they blend by computing either a simple average (bi-SA, Eq.~\ref{eq:bi-sa}) or a weighted average based on the scaled time step $w_t$ (bi-TW, Eq.~\ref{eq:bi-tw}):
\begin{align}
\widehat{v}_t &= \left(\widehat{v}_t^P + \widehat{v}_t^F\right)/2 \, , \label{eq:bi-sa} \\
\widehat{v}_t &= \left( 1-w_t \right) \widehat{v}_t^P + w_t \widehat{v}_t^F \, . \label{eq:bi-tw}
\end{align}
All baselines are trained independently from scratch.

\subsection{KTH Actions}
\label{sec:results_kth_actions}

\begin{figure}[t]
	\centering
    \includegraphics[width=\textwidth]{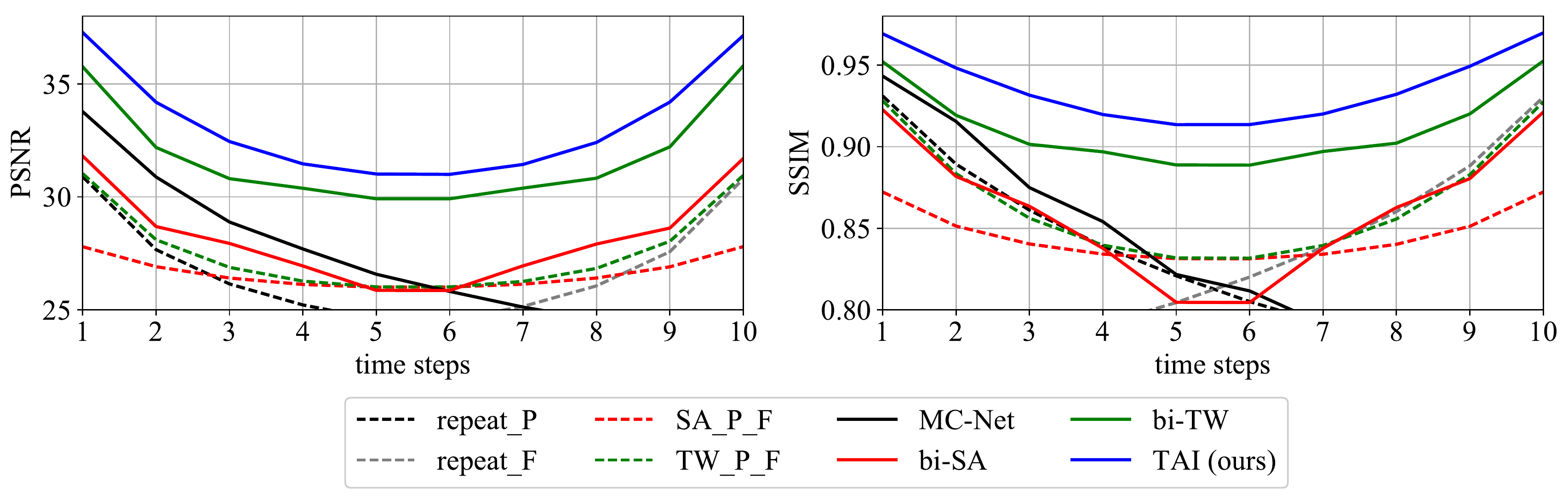}
    \caption{Quantitative results on the KTH Actions dataset (for both PSNR and SSIM, higher is better). We compare our full model (TAI) to the baselines described in Sec.~\ref{sec:baselines}}
    \label{fig:kth_quant}
\end{figure}

\begin{figure}[t]
    \centering
    \includegraphics[width=\textwidth]{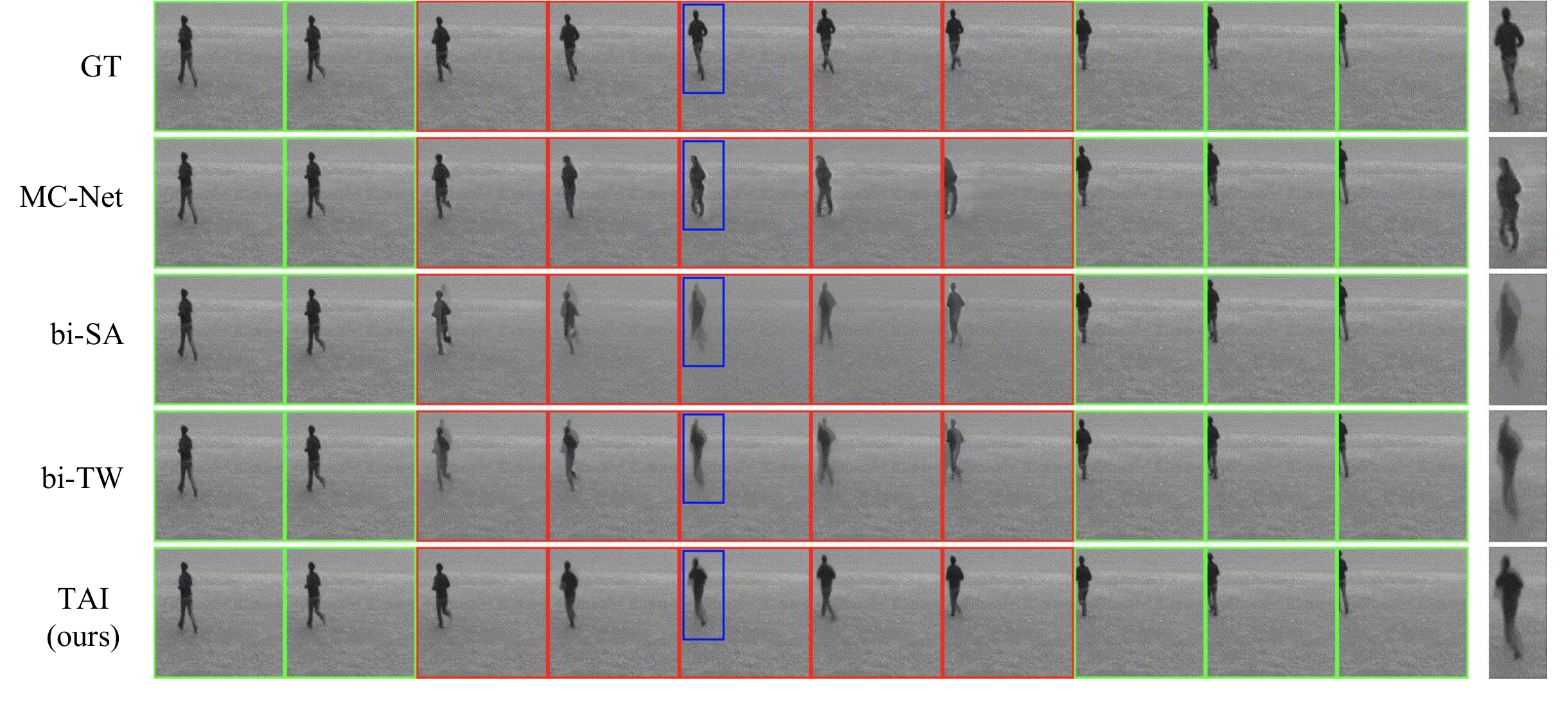}
    \caption{Comparison of predictions from our TAI model to baseline methods on KTH Actions. We visualize every other frame of the input and predicted sequences. Refer to the supplementary materials for more results}
    \label{fig:kth_qual}
\end{figure}

To evaluate the performance of the proposed baselines and our full model with the TAI network (we refer to this full model as TAI for brevity), we follow existing video prediction literature~\cite{villegas2017decomposing,mathieu2015deep} by reporting the Structural Similarity (SSIM)~\cite{wang2004image} and Peak Signal-Noise Ratio (PSNR) between each predicted frame and the ground truth. We draw a series of conclusions from the quantitative comparison shown in Fig.~\ref{fig:kth_quant}. First, the low performance of the hand-crafted baselines (the dashed curves in Fig.~\ref{fig:kth_quant}) indicate that our task is challenging, and requires a model that generates a \textit{non-trivial} prediction from \textit{multiple} preceding and following frames. Second, the performance of MC-Net drops quickly over time due to its lack of guidance from the following frames. Third, between the bidirectional prediction baselines, bi-TW does a better job than bi-SA since it incorporates the scaled time step $w_t$ via a hand-crafted, time-weighted average. Finally, TAI outperforms bi-TW because it learns a complex blending function that leverages both time step information and intermediate activations from the Bidirectional Video Prediction Network.

In Fig.~\ref{fig:kth_qual}, we visualize the predictions made by MC-Net, bi-SA, bi-TW, and TAI (we encourage the reader to view additional results in the supplementary materials). MC-Net generates blob-like poses that fail to preserve the proper shape of the body and are inconsistent with the following frames. Meanwhile, bi-SA and bi-TW generate frames with a noticeable ``ghosting'' effect (e.g. both predictions contain two torsos overlapping with each other), leading to a drop in PSNR and SSIM scores. On the other hand, TAI overcomes these challenges: its predictions are consistent with both the preceding and following frames, and they contain one unified, well-shaped torso. We have found that SSIM drops more drastically than PSNR when ghosting occurs, suggesting that it correlates better with human-perceived quality than PSNR.

\subsection{Qualitative Analysis of Blending Methods}

\begin{figure}[t]
	\centering
    \includegraphics[width=0.9\textwidth]{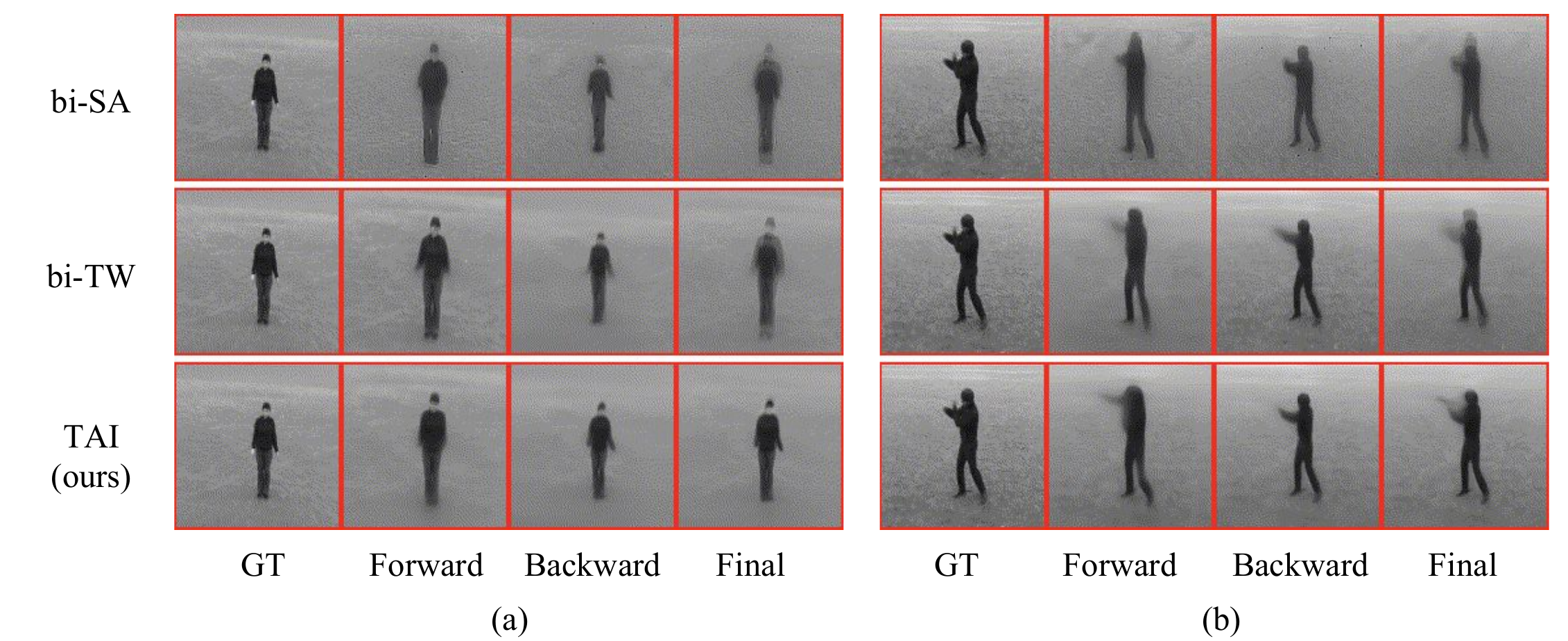}
    \caption{Comparison of the forward, backward, and final predictions for the third middle frame (of ten) of two videos}
   	\label{fig:intermediate_results}
\end{figure}

Next, we visualize the forward, backward, and final predictions of bi-SA, bi-TW, and TAI in order to highlight the benefit of a learned blending function over a hand-crafted one. Across all three models, the forward prediction is inconsistent with the backward one for most videos. For instance, in Fig.~\ref{fig:intermediate_results}, the scale of the actor always differs between the forward and backward predictions. However, the quality of the final prediction improves with the complexity of the blending strategy. For example, since bi-SA blends the two predictions evenly, we observe in the final prediction for Fig.~\ref{fig:intermediate_results}a a blurry background and two outlines of the actor's body; in Fig.~\ref{fig:intermediate_results}b, we see the outlines of two heads. bi-TW produces similar artifacts to bi-SA for both videos, but its final predictions are clearer. Finally, TAI reconciles the differences between the forward and backward predictions without introducing ghosting artifacts: in Fig.~\ref{fig:intermediate_results}a, the final prediction compromises between the actor's sizes from the intermediate predictions, and in Fig.~\ref{fig:intermediate_results}b, the difference in the actor's head position is resolved, resulting in a clean outline of the head. We conclude that even though all three methods generate inconsistent forward and backward predictions, TAI can successfully reconcile the differences to generate a crisp final prediction.

\subsection{Ablation Studies}
\label{sec:ablation_studies}

\begin{figure}[t]
	\begin{subfigure}[b]{0.4\textwidth}
		\centering
        \includegraphics[width=\textwidth]{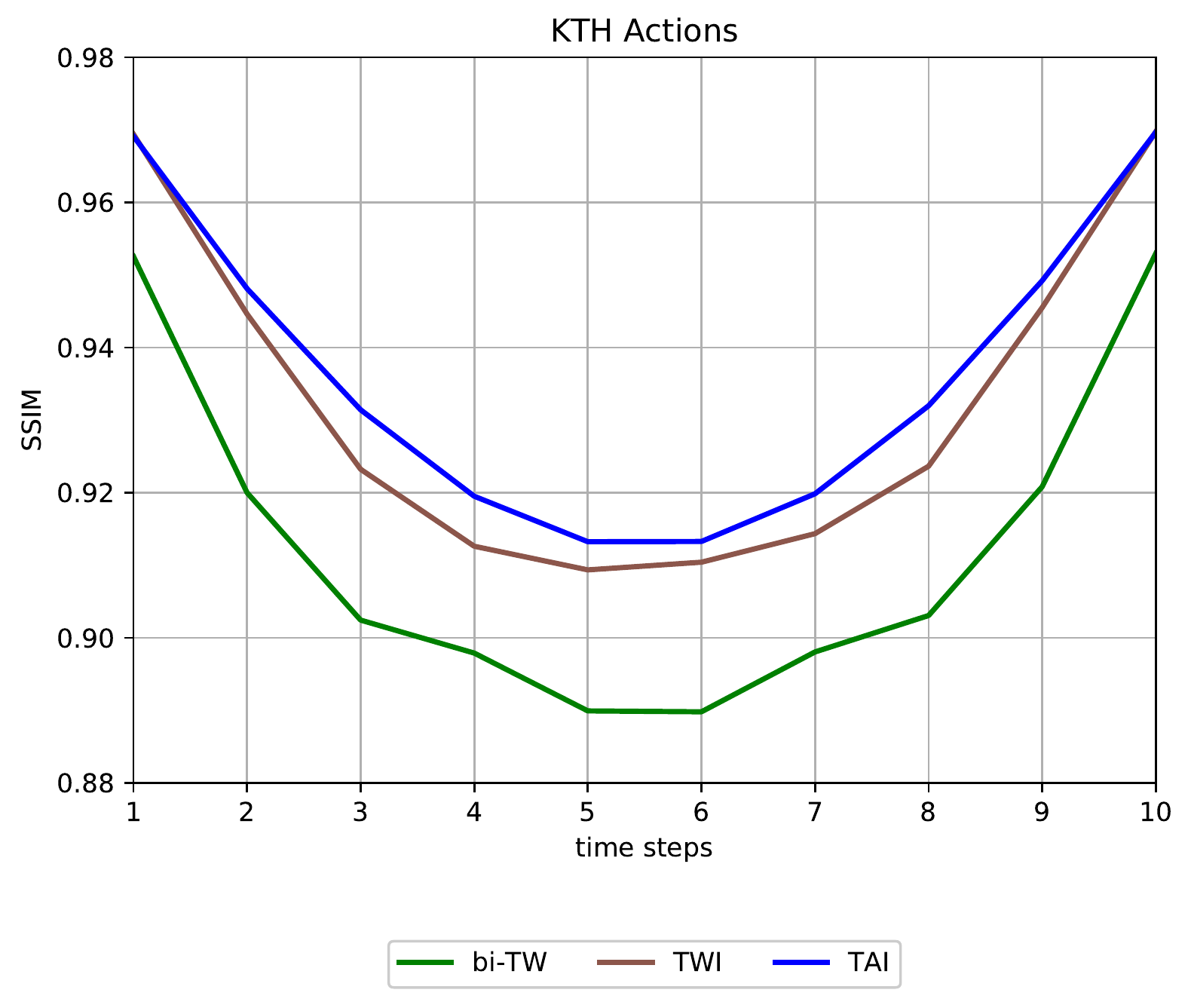}
        \caption{SSIM}
        \label{fig:ablation_ssim}
	\end{subfigure}
	\begin{subfigure}[b]{0.6\textwidth}
		\centering
        \includegraphics[width=\textwidth]{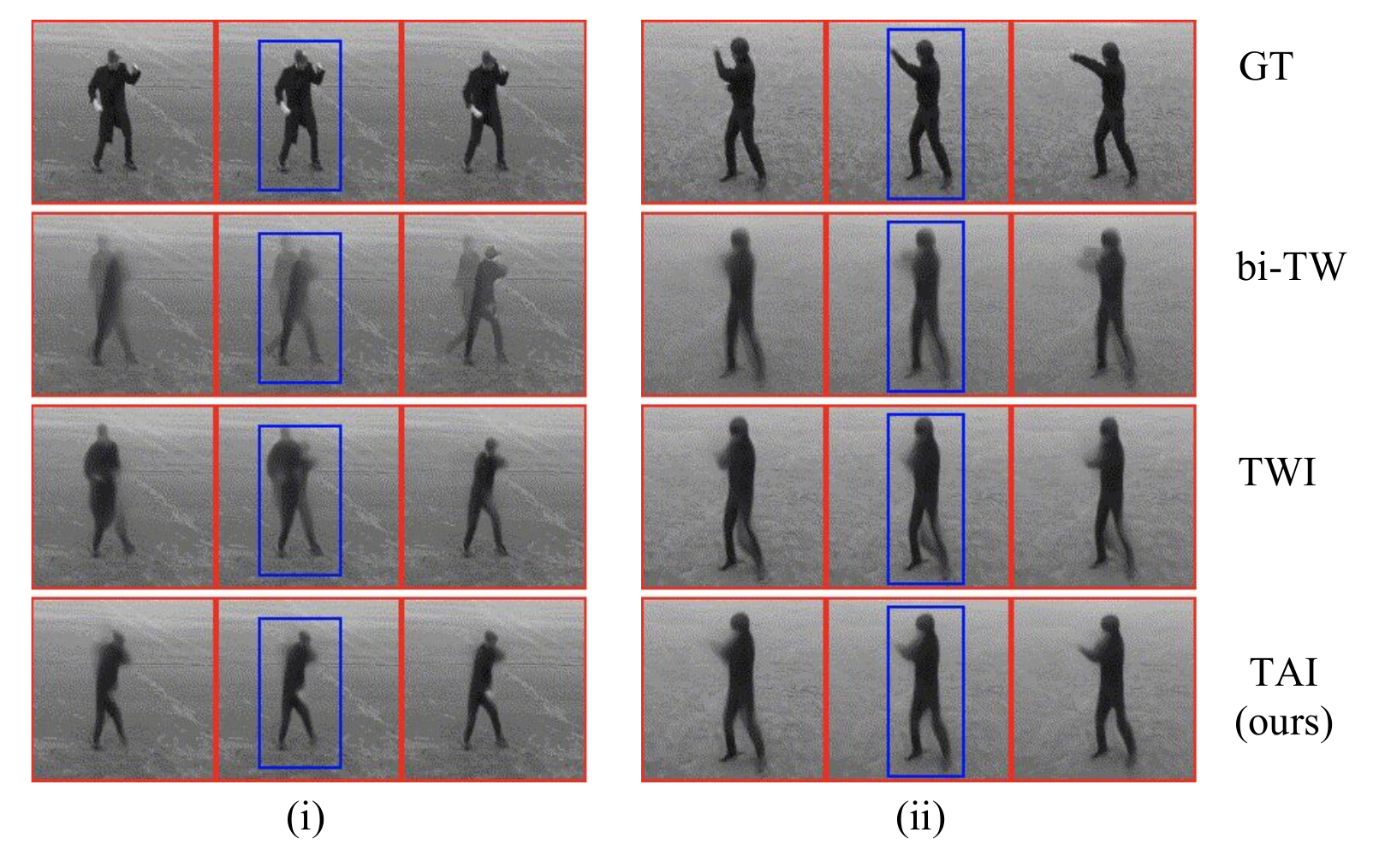}
        \caption{Example videos}
        \label{fig:ablation_sample}
	\end{subfigure}
    \caption{Ablative comparison between bi-TW, TWI, and our full TAI model. Higher SSIM is better}
    \label{fig:itwi_vs_itai}
\end{figure}

Feeding time information into the blending network such that it can \textit{learn} to use that information most effectively is key to generating high-quality predictions. To verify this, we replace the blending module with a time-agnostic interpolation network and apply a time-weighted average to its outputs; we call this version the time-weighted interpolation (TWI) network. In Fig.~\ref{fig:itwi_vs_itai}, we compare bi-TW, TAI, and a bidirectional video prediction model with TWI. We see that TWI performs better than bi-TW both quantitatively and qualitatively because the ghosting artifacts in its predictions are less apparent. However, it still incorporates time information with a hand-crafted function, which prevents TWI from completely avoiding ghosting artifacts. For example, TWI generates two torsos in Fig.~\ref{fig:ablation_sample}~(i) and a fake leg between two legs in Fig.~\ref{fig:ablation_sample}~(ii).
On the other hand, TAI avoids ghosting artifacts more successfully than TWI: for both videos in Fig.~\ref{fig:ablation_sample}, we see that TAI generates a clear, sharp outline around the actor without introducing artificial torsos or limbs.

\subsection{Importance of Context Frames}

\begin{figure}[t]
	\centering
    \includegraphics[width=\textwidth]{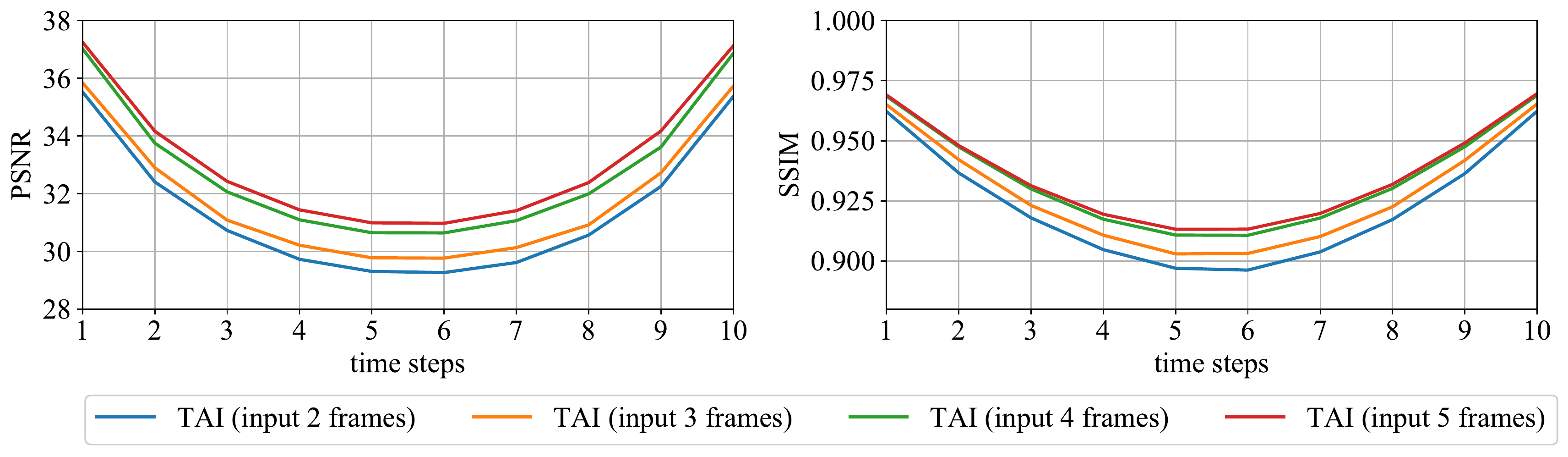}
    \caption{Performance of our trained model with 2-5 preceding and following frames at test time on the KTH Actions dataset. Higher PSNR and SSIM is better}
    \label{fig:supp_context}
\end{figure}

In this section, we show our model's ability to leverage the context information from the preceding and the following sequences which, as argued in Sec.~\ref{sec:introduction}, is vital to performing well on the video frame inpainting task. In Fig.~\ref{fig:supp_context}, we plot the quantitative performance of our trained model as we increase the number of available frames at test time from two to five (recall that we train our model on five preceding and following frames). Our model obtains progressively better PSNR and SSIM values as we increase the number of preceding and following frames; this shows that our model successfully leverages the increasing amount of context information to improve its predictions.

\begin{figure}[t]
	\centering
    \includegraphics[width=\textwidth]{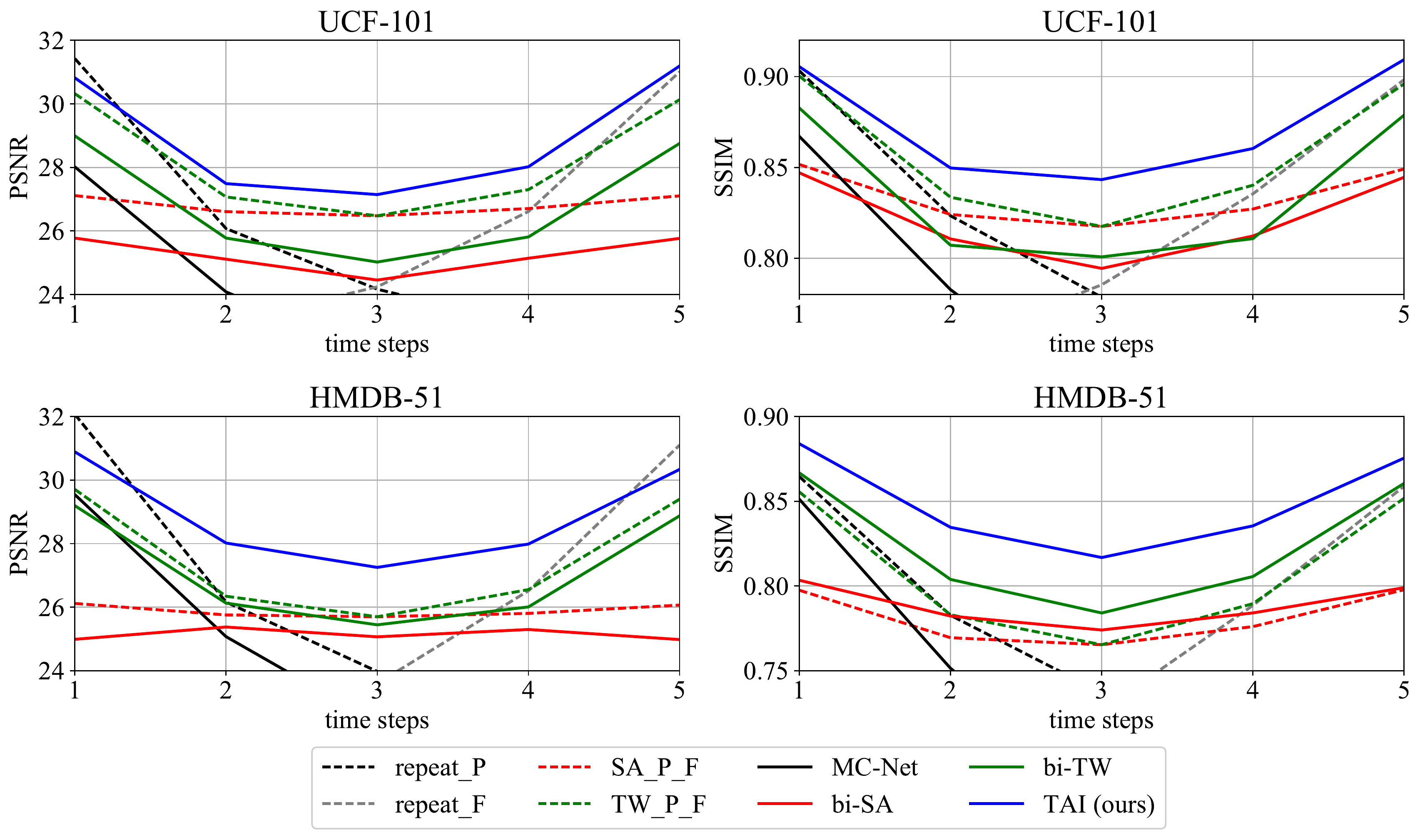}
    \caption{Quantitative results on the UCF-101 and HMDB-51 datasets (higher PSNR and SSIM is better). We compare our method to the baselines described in Sec.~\ref{sec:baselines}}
    \label{fig:ucf_hmdb_quant}
\end{figure}

\begin{figure}[t]
    \centering
    \begin{subfigure}[b]{\textwidth}
    \centering
        \includegraphics[width=0.9\textwidth]{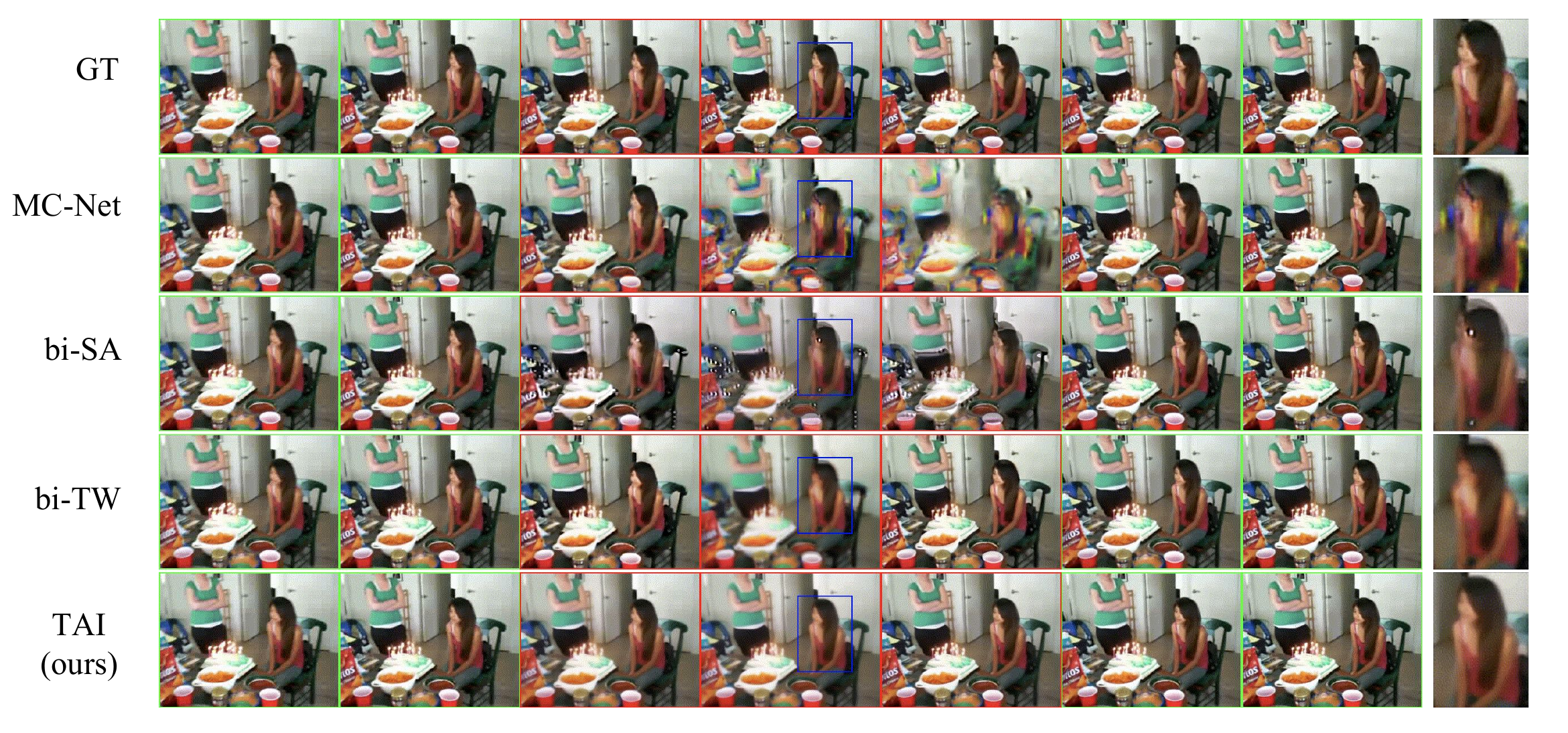}
        \caption{UCF-101}
        \label{fig:ucf_qual}
    \end{subfigure}
    \begin{subfigure}[b]{\textwidth}
    \centering
        \includegraphics[width=0.9\textwidth]{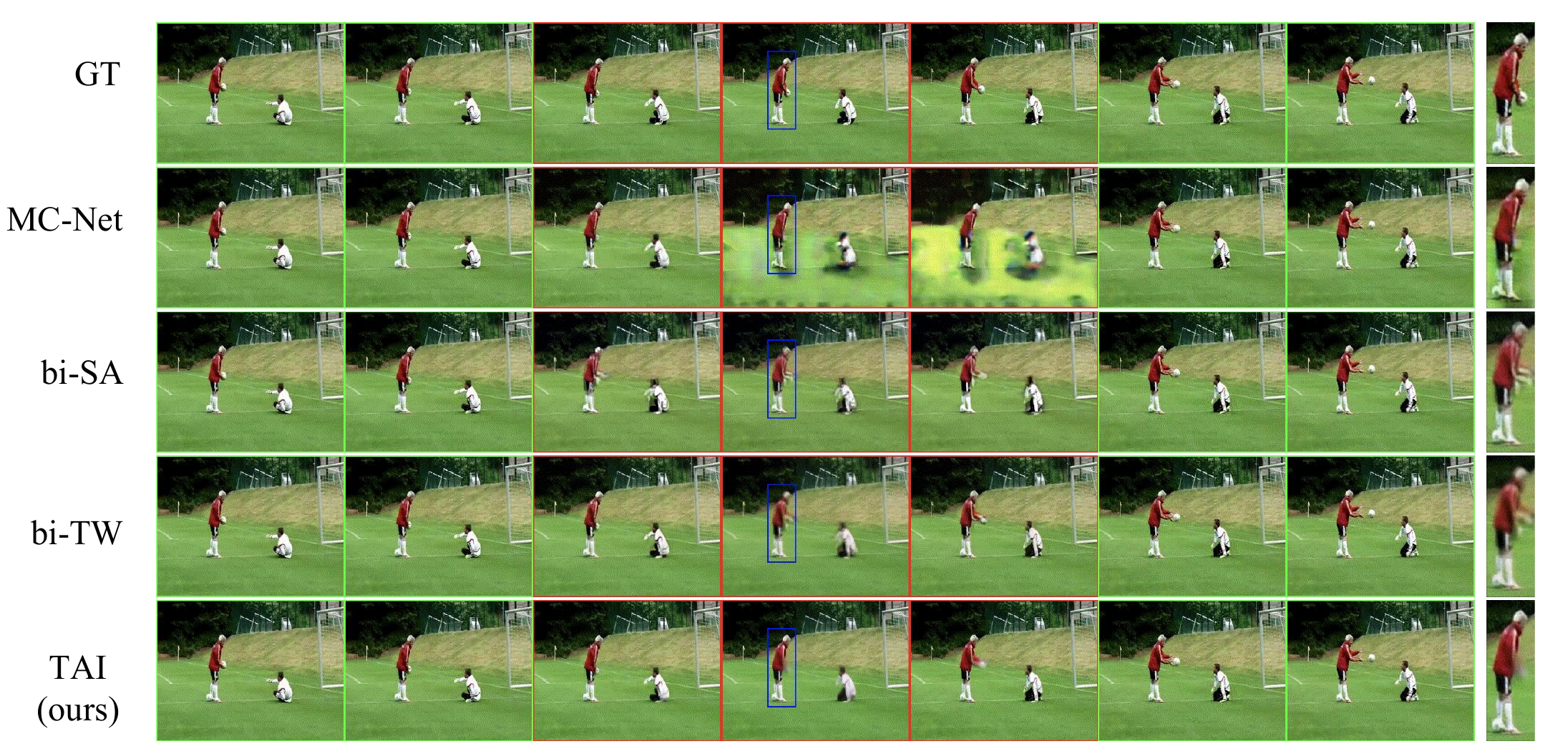}
        \caption{HMDB-51}
        \label{fig:hmdb_qual}
    \end{subfigure}
    \caption{Comparison of predictions from our approach to baseline methods on the UCF-101 and HMDB-51 datasets. We visualize every second frame of the input and predicted sequences. Refer to the supplementary materials for more results}
    \label{fig:ucf_hmdb_qual}
\end{figure}

\subsection{HMDB-51 and UCF-101}

We conclude our experiments by demonstrating our model's ability to perform well on complex videos depicting challenging scenes and a wide variety of actions. We do this by comparing our full TAI model to the baselines proposed in Sec.~\ref{sec:baselines} on videos from HMDB-51 and UCF-101.
We see from the quantitative results in Fig.~\ref{fig:ucf_hmdb_quant} that none of the baselines outperform the others by a definitive margin. This contrasts with our findings in Sec.~\ref{sec:results_kth_actions} where we found that for KTH Actions, bi-TW produces substantially better predictions than all the other baselines. We note that the biggest difference between KTH Actions and HMDB-51/UCF-101 is that the scenes in HMDB-51 and UCF-101 are far more complex than in KTH Actions; this suggests that bi-TW performs poorly when observing complex scenes. However, our model still outperforms all baselines on HMDB-51 and UCF-101, suggesting that it is best equipped for handling complex videos.

We present qualitative comparisons in Fig.~\ref{fig:ucf_hmdb_qual}. In Fig.~\ref{fig:ucf_qual}, we observe two contours of the girl's hair in the bi-SA prediction, and a blurry facial expression in the bi-TW prediction. On the other hand, our TAI model generates a unified contour of the hair and a clear facial expression. Moving on to Fig.~\ref{fig:hmdb_qual}, we note that MC-Net distorts the background in the later middle frames, and that both bi-SA and bi-TW generate blurry patterns on the man's jacket and pants. However, TAI produces a clear white stripe on the man's pants, as well as a sharp outline around his jacket. Our results demonstrate that on video datasets containing complex scenes and a large number of action classes, TAI generates predictions that are more visually satisfying than several strong baselines.

\section{Conclusion}

In this paper, we have tackled the video frame inpainting problem by generating two sets of intermediate predictions conditioned on the preceding and following frames respectively, and then blending them together with our novel TAI network. Our experiments on KTH Actions, HMDB-51, and UCF-101 show that our method generates more accurate and visually pleasing predictions than multiple strong baselines. Furthermore, our in-depth analysis has revealed that our TAI network successfully leverages time step information to reconcile inconsistencies in the intermediate predictions, and that it leverages the full context provided by the preceding and following frames.
In future work, we aim to improve performance by exploiting semantic knowledge about the video content, e.g. by modeling human poses or the periodicity of certain actions. We also aim to explore models that can predict an even greater number of frames, i.e. several seconds of video instead of fractions of a second. To encourage innovations in deep learning for video frame inpainting, we have made our code publicly available at \url{https://github.com/sunxm2357/TAI_video_frame_inpainting}.

\section{Acknowledgements}

This work is partly supported by ARO W911NF-15-1-0354, DARPA FA8750-17-2-0112 and DARPA FA8750-16-C-0168. It reflects the opinions and conclusions of its authors, but not the funding agents.

\newpage

\appendix

\vspace*{0cm}

\begin{center}
    \textbf{\Large{A Temporally-Aware Interpolation Network for}}
    
    \textbf{\Large{Video Frame Inpainting:}}
    
    \textbf{\Large{Supplementary Materials}}
\end{center}

\section{Qualitative Results on KTH Actions}
In this section, we give six more qualitative results on the KTH Actions dataset, one for each action class. To save space, we visualize every other frame. Our full model TAI gives more visually pleasing predictions on all six classes than MC-Net, bi-SA and bi-TW.

\begin{figure}[H]
    \vspace{-10pt}
    \centering
    \includegraphics[width=\textwidth]{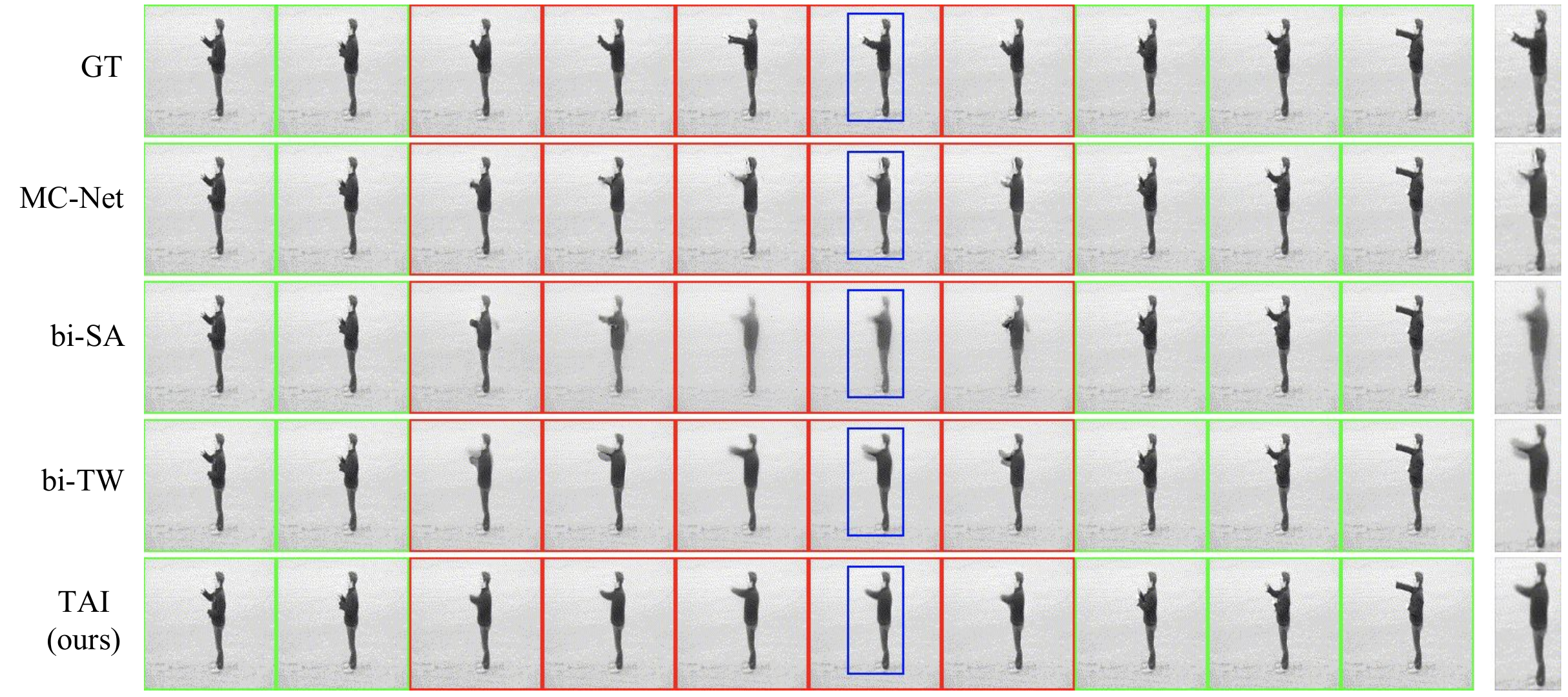}
    \vspace{-15pt}
    \caption{Comparison on a ``boxing'' video from KTH Actions}
    \label{fig:kth_boxing}
    \vspace{-20pt}
\end{figure}
\begin{figure}[H]
    \centering
    \vspace{-15pt}
    \includegraphics[width=\textwidth]{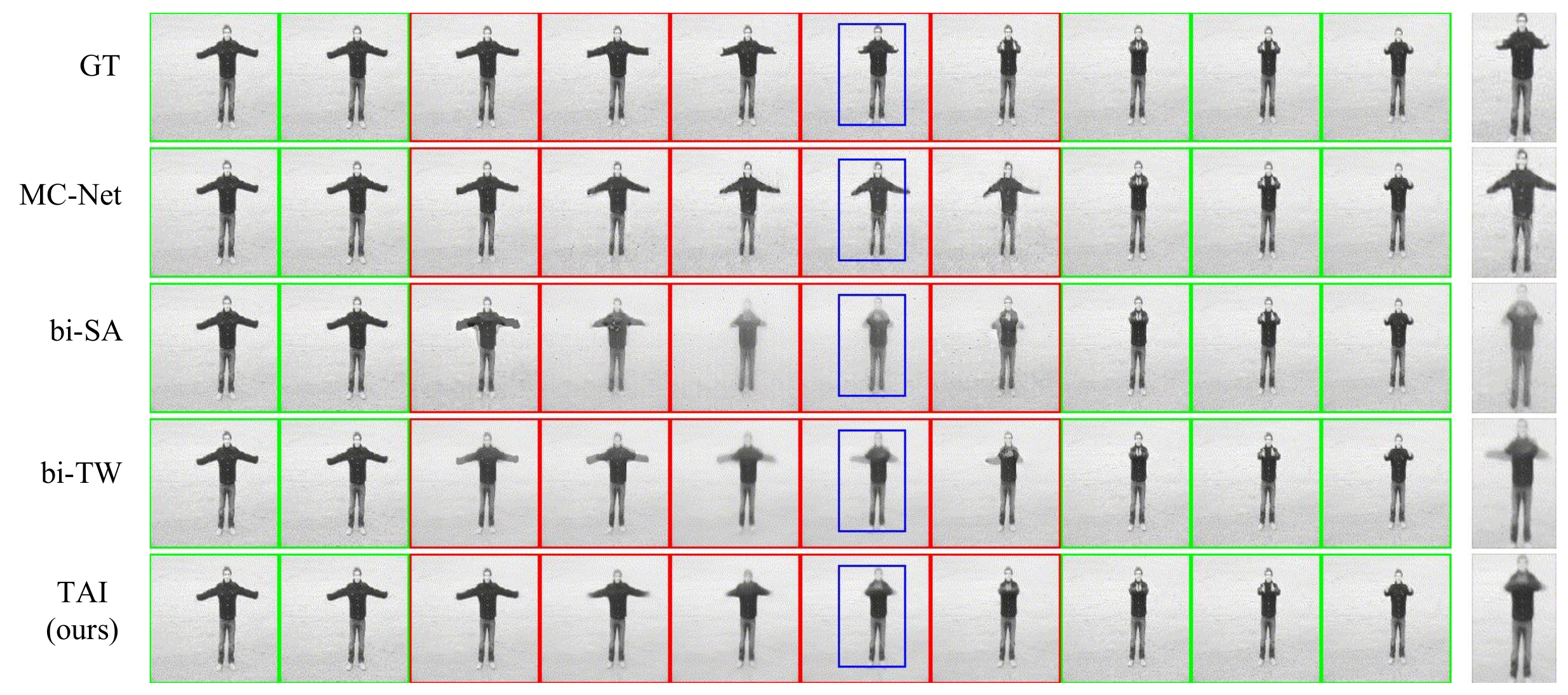}
    \vspace{-15pt}
    \caption{Comparison on a ``handclapping'' video from KTH Actions}
    \label{fig:kth_handclapping}
    \vspace{-40pt}
\end{figure}
\begin{figure}[H]
    \centering
    \includegraphics[width=\textwidth]{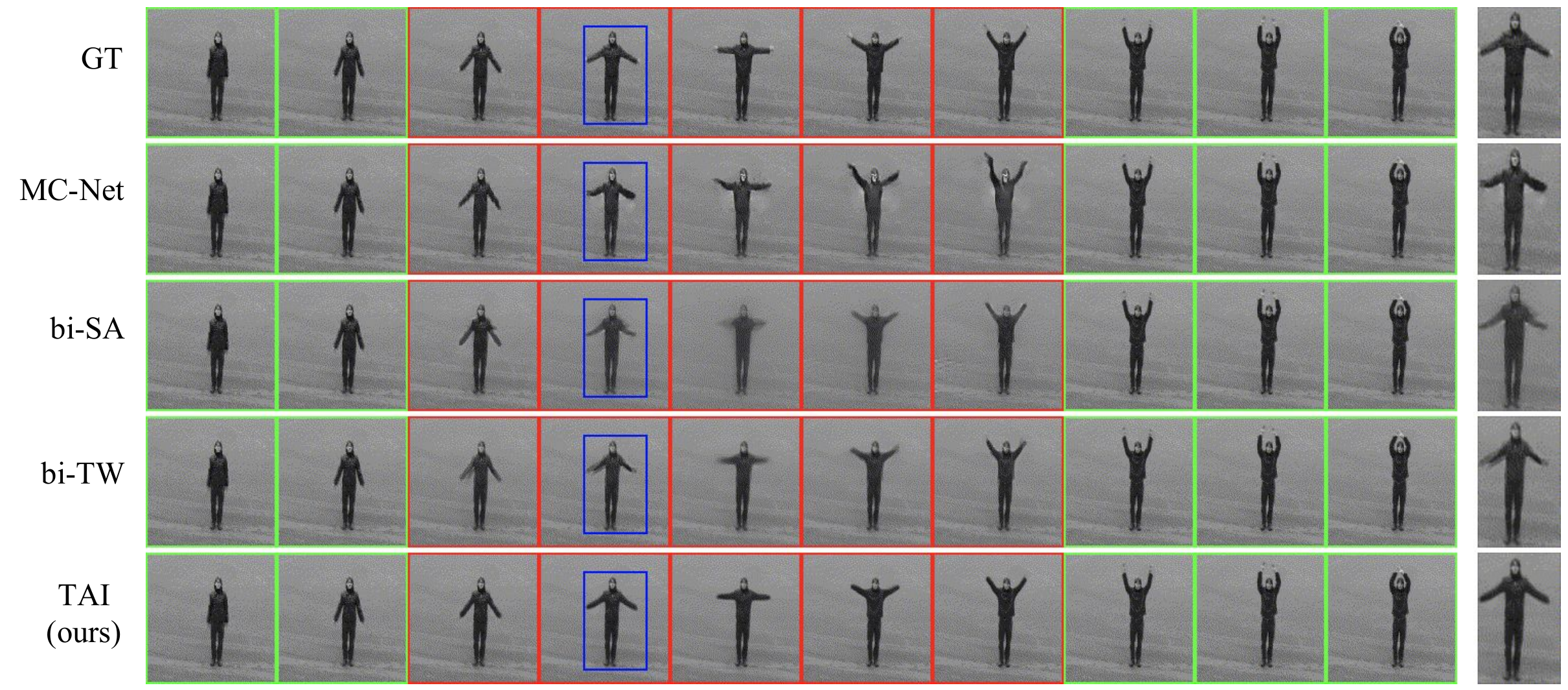}
    \vspace{-15pt}
    \caption{Comparison on a ``handwaving'' video from KTH Actions}
    \label{fig:kth_handwaving}
    \vspace{-40pt}
\end{figure}
\begin{figure}[H]
    \centering
    \includegraphics[width=\textwidth]{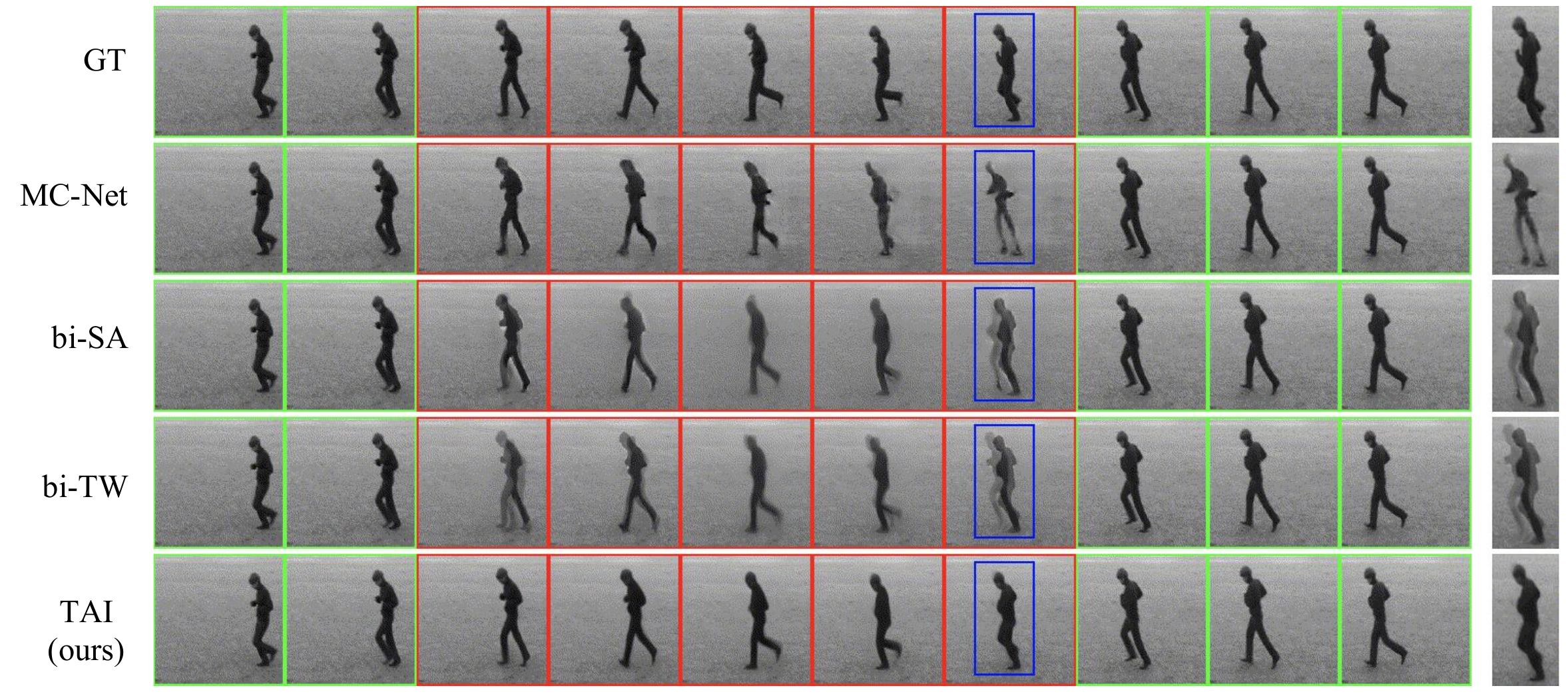}
    \vspace{-15pt}
    \caption{Comparison on a ``jogging'' video from KTH Actions}
    \label{fig:kth_jogging}
    \vspace{-40pt}
\end{figure}
\begin{figure}[H]
    \centering
    \includegraphics[width=\textwidth]{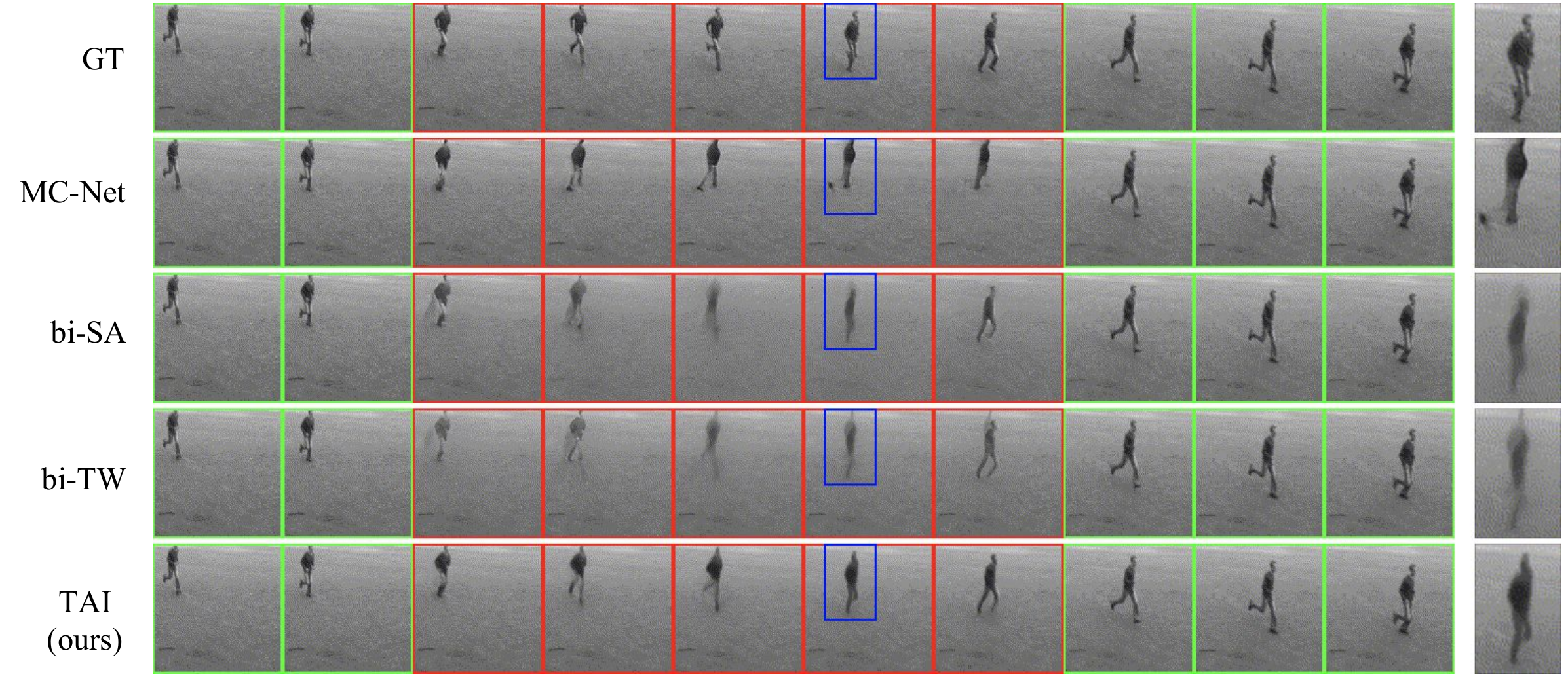}
    \vspace{-15pt}
    \caption{Comparison on a ``running'' video from KTH Actions}
    \label{fig:kth_running}
\end{figure}
\begin{figure}[H]
    \centering
    \includegraphics[width=\textwidth]{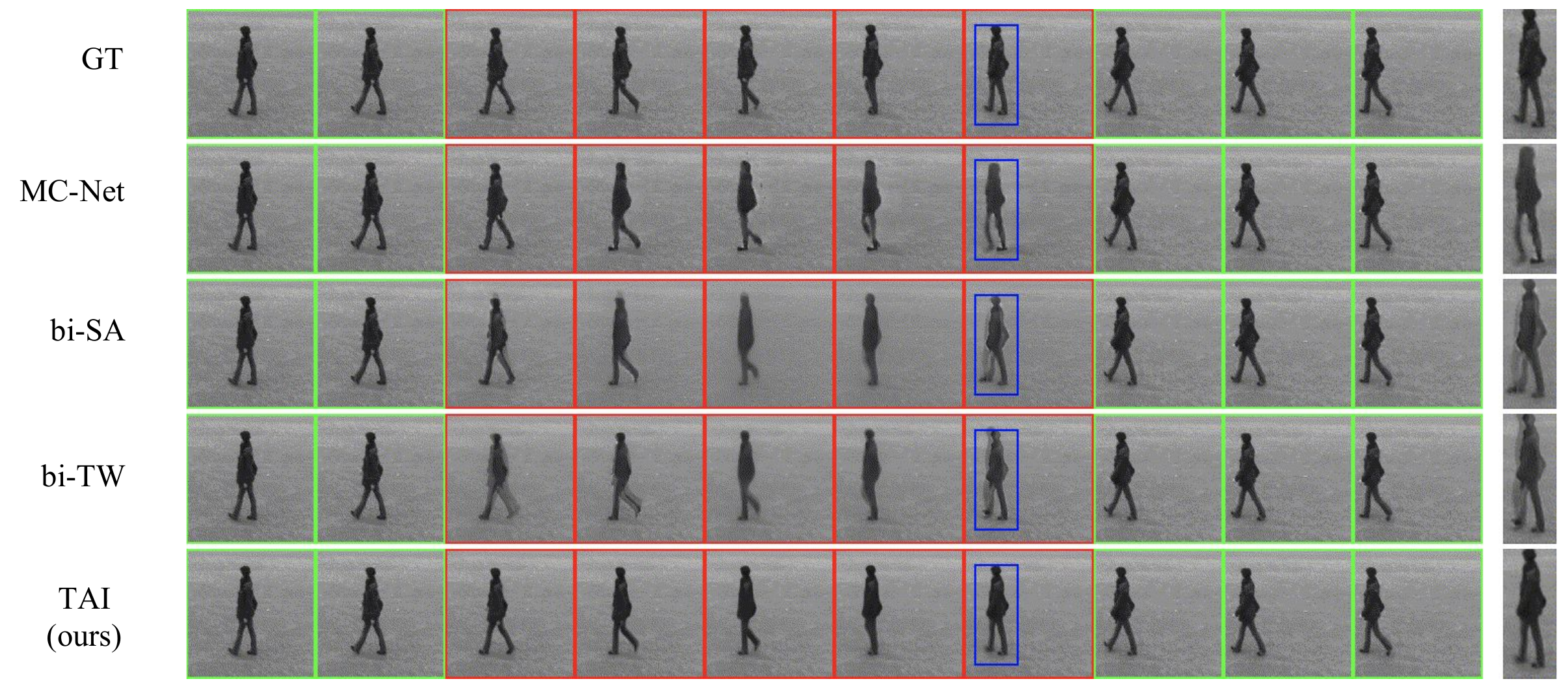}
    \vspace{-15pt}
    \caption{Comparison on a ``walking'' video from KTH Actions}
    \label{fig:kth_walking}
\end{figure}

\section{Qualitative Results on UCF-101 and HMDB-51}
In this section, we give six more qualitative results on different action classes, three from UCF-101 and three from HMDB-51. To save space, we visualize every other frame. Our full model TAI reconciles the misalignment of the forward and backward predictions and gives a crisp prediction.

\begin{figure}[H]
    \centering
    \includegraphics[width=\textwidth]{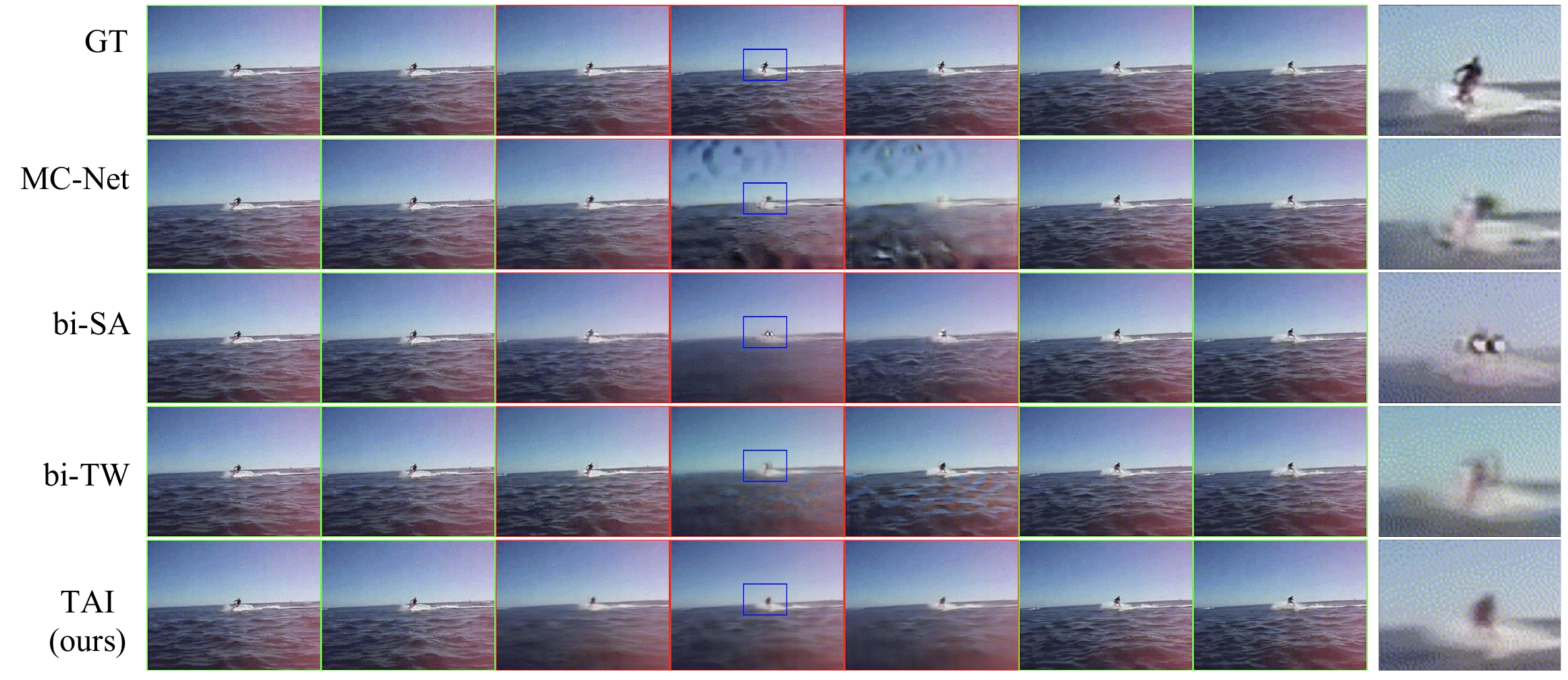}
    \vspace{-15pt}
    \caption{Comparison on a ``Skijet'' video from UCF-101}
    \label{fig:ucf_Skijet}
    \vspace{-20pt}
\end{figure}

\newpage

\begin{figure}[H]
    \centering
    \includegraphics[width=\textwidth]{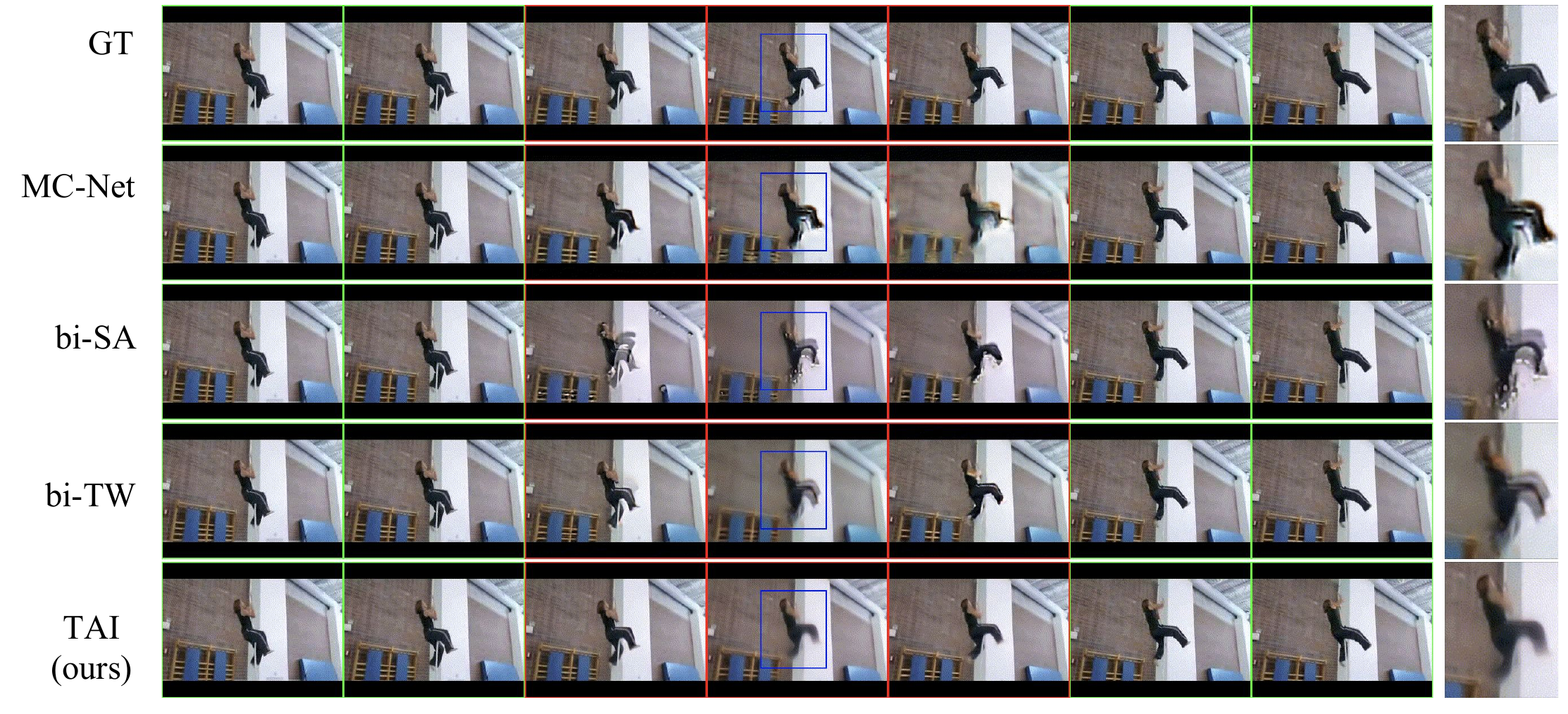}
    \vspace{-15pt}
    \caption{Comparison on a ``RopeClimbing'' video from UCF-101}
    \label{fig:ucf_RopeClimbing}
    \vspace{-40pt}
\end{figure}

\begin{figure}[H]
    \centering
    \includegraphics[width=\textwidth]{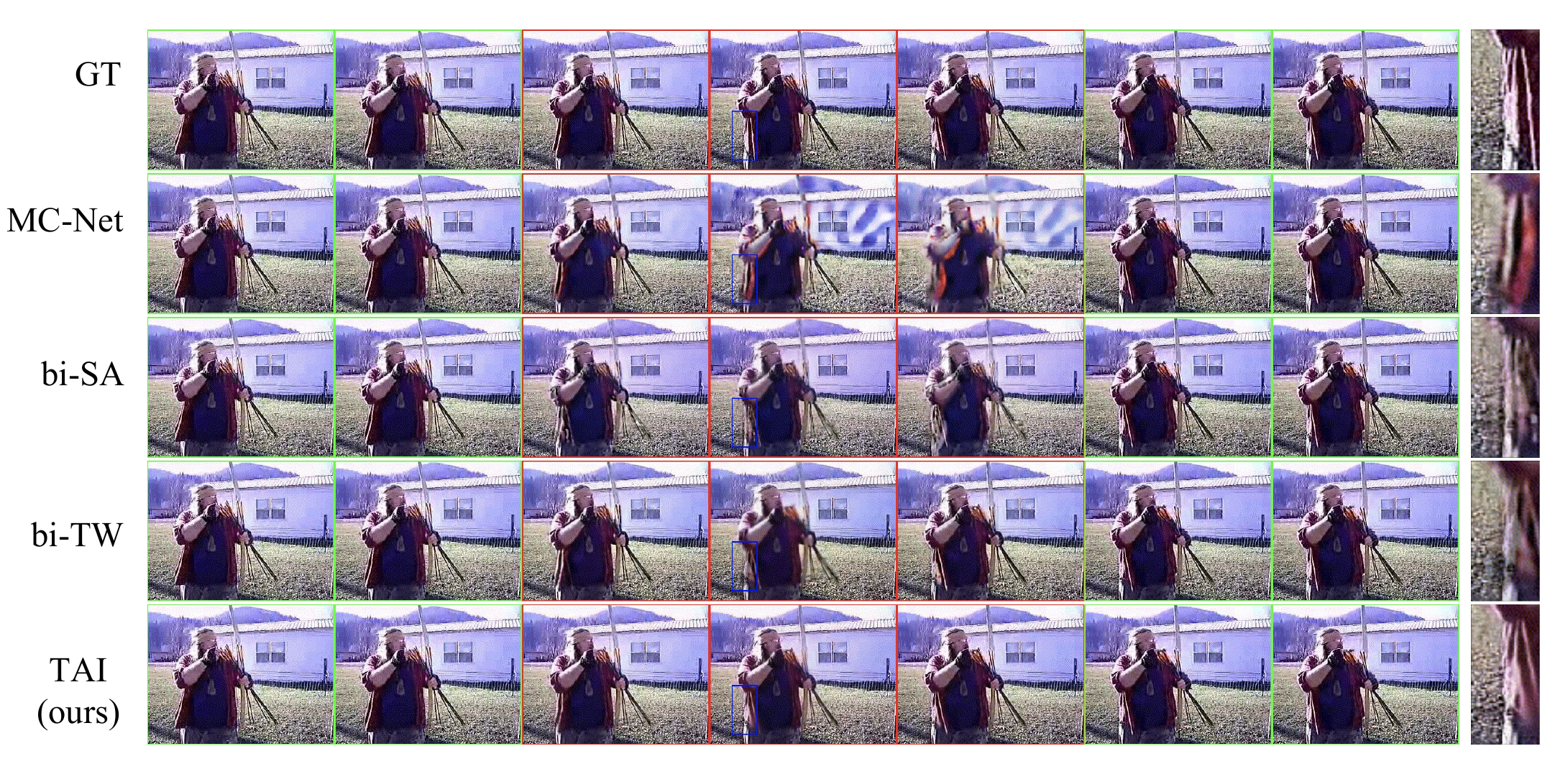}
    \vspace{-15pt}
    \caption{Comparison on an ``Archery'' video from UCF-101}
    \label{fig:ucf_Archery}
    \vspace{-40pt}
\end{figure}

\begin{figure}[H]
    \centering
    \includegraphics[width=\textwidth]{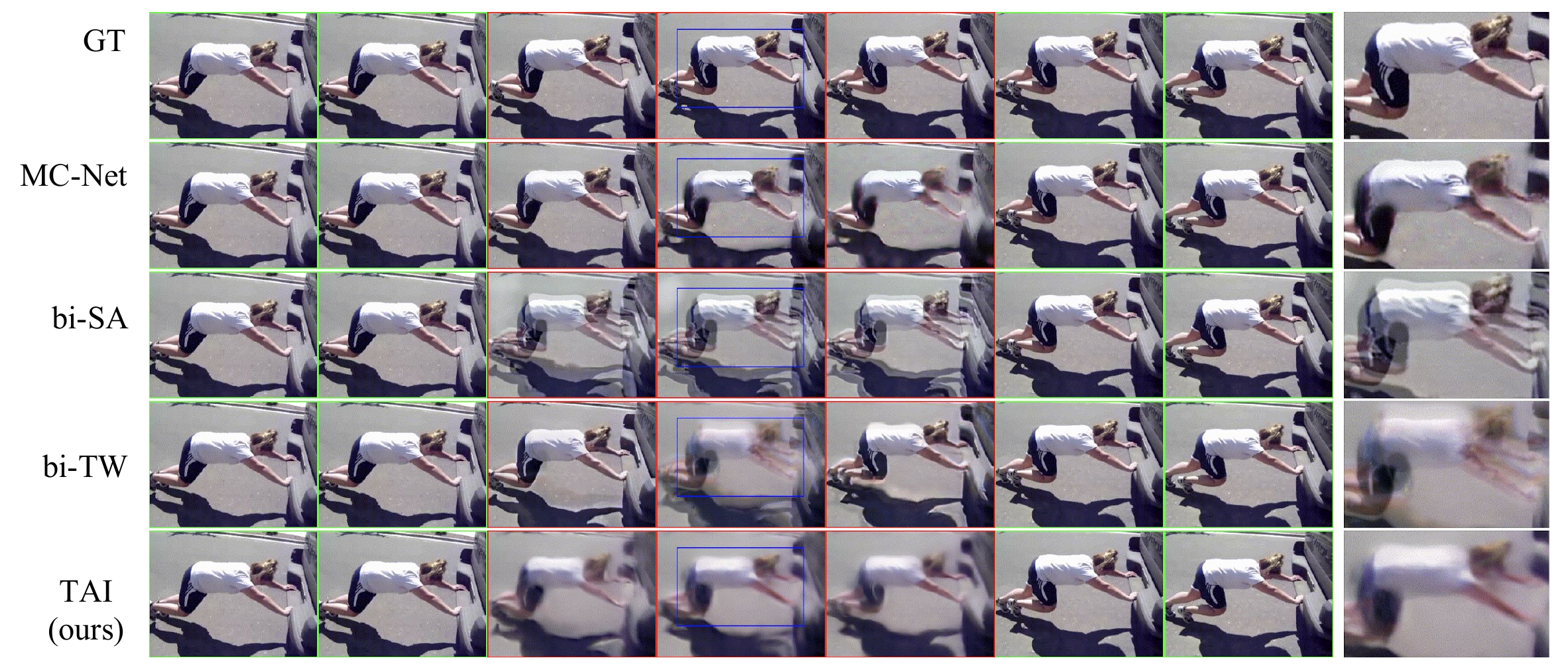}
    \vspace{-15pt}
    \caption{Comparison on a ``push'' video from HMDB-51}
    \label{fig:hmdb_push}
\end{figure}

\begin{figure}[H]
    \centering
    \includegraphics[width=\textwidth]{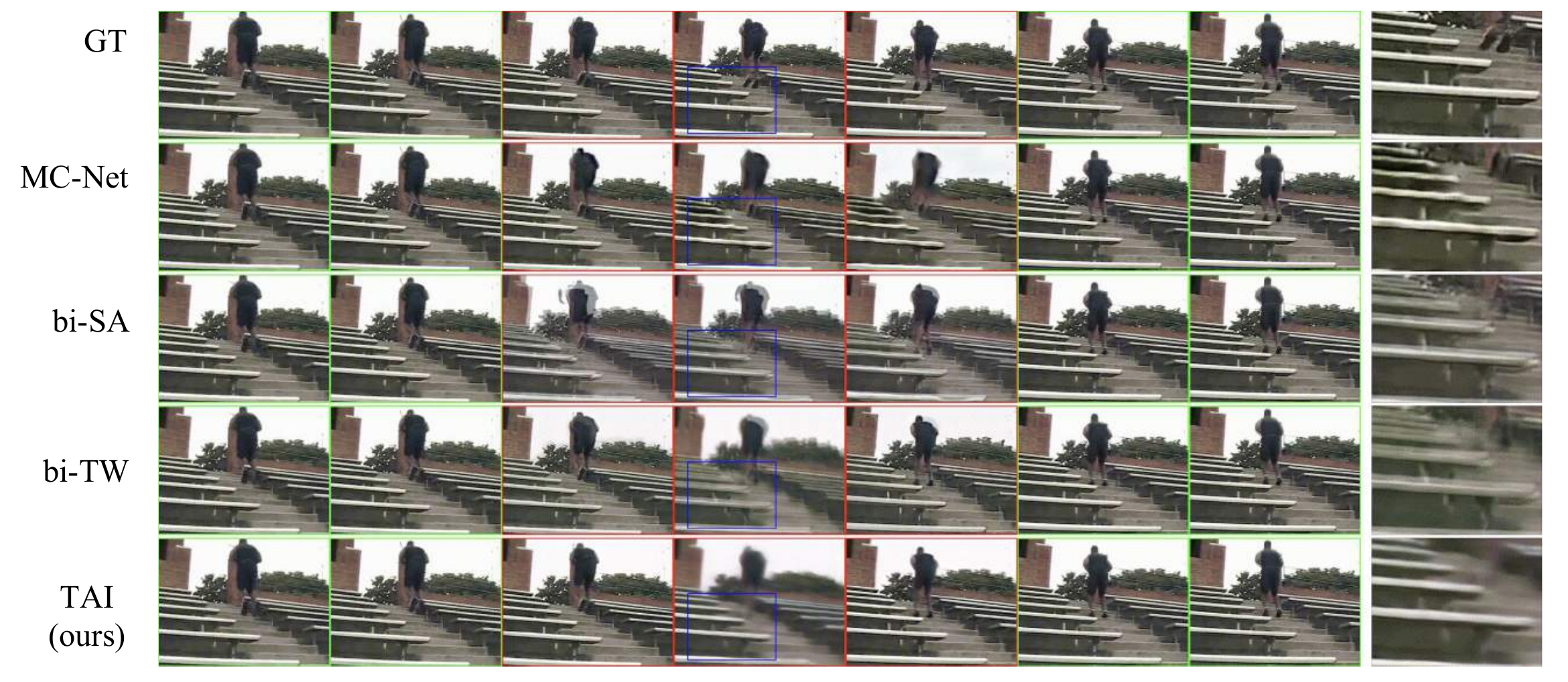}
    \vspace{-15pt}
    \caption{Comparison on a ``jump'' video from HMDB-51}
    \label{fig:hmdb_jump}
    \vspace{-40pt}
\end{figure}

\begin{figure}[H]
    \centering
    \includegraphics[width=\textwidth]{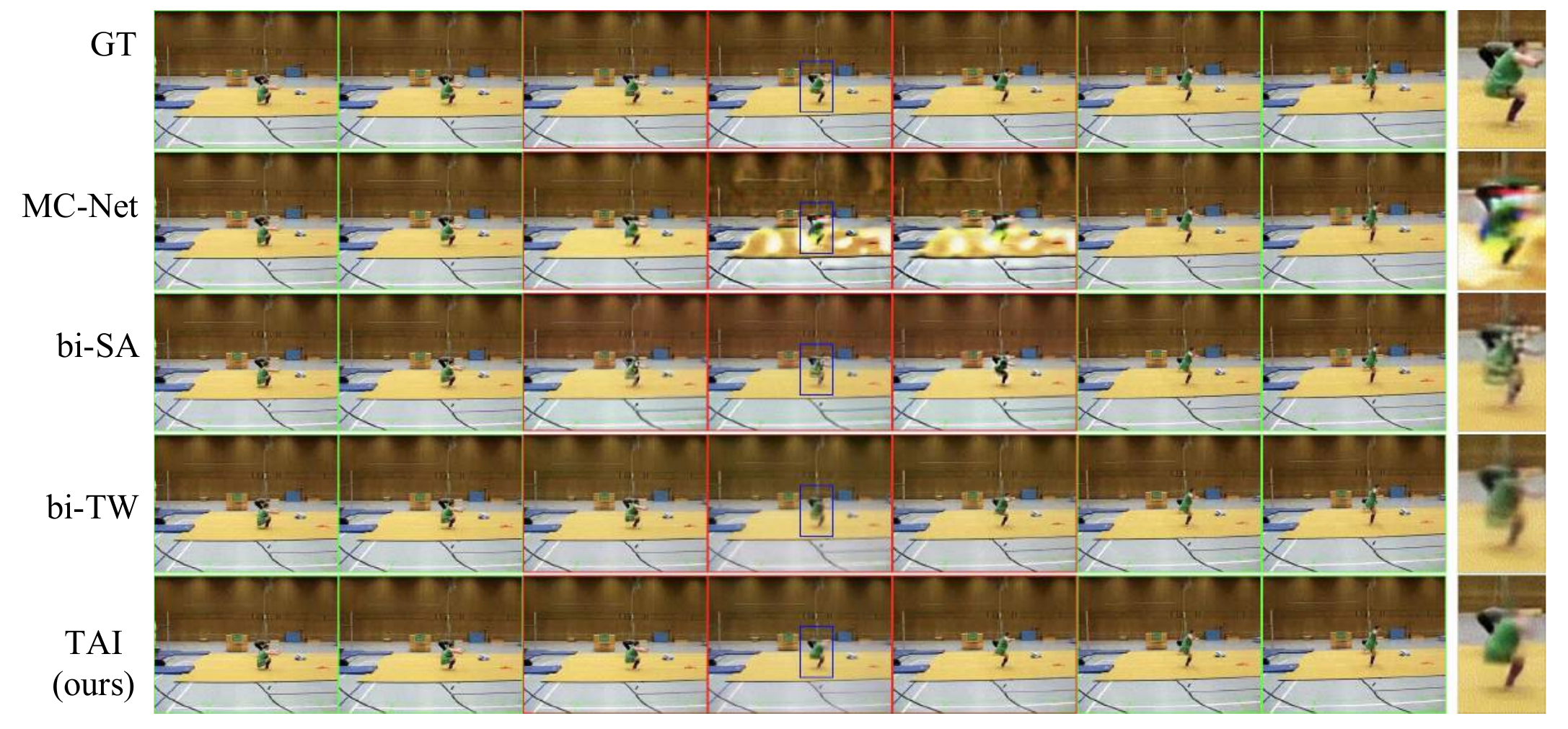}
    \vspace{-15pt}
    \caption{Comparison on a ``somersault'' video from HMDB-51}
    \label{fig:hmdb_somersault}
\end{figure}


\section{Importance of Context Frames: Qualitative Analysis}

In addition to the quantitative results in the main paper, we provide the qualitative results on the KTH Actions dataset to demonstrate that our TAI model can leverage the context information from the preceding and following frames to better predict the middle frames.

\begin{figure}[H]
	\centering
    \vspace{-20pt}
    \begin{subfigure}[b]{\textwidth}
        \includegraphics[width=\textwidth]{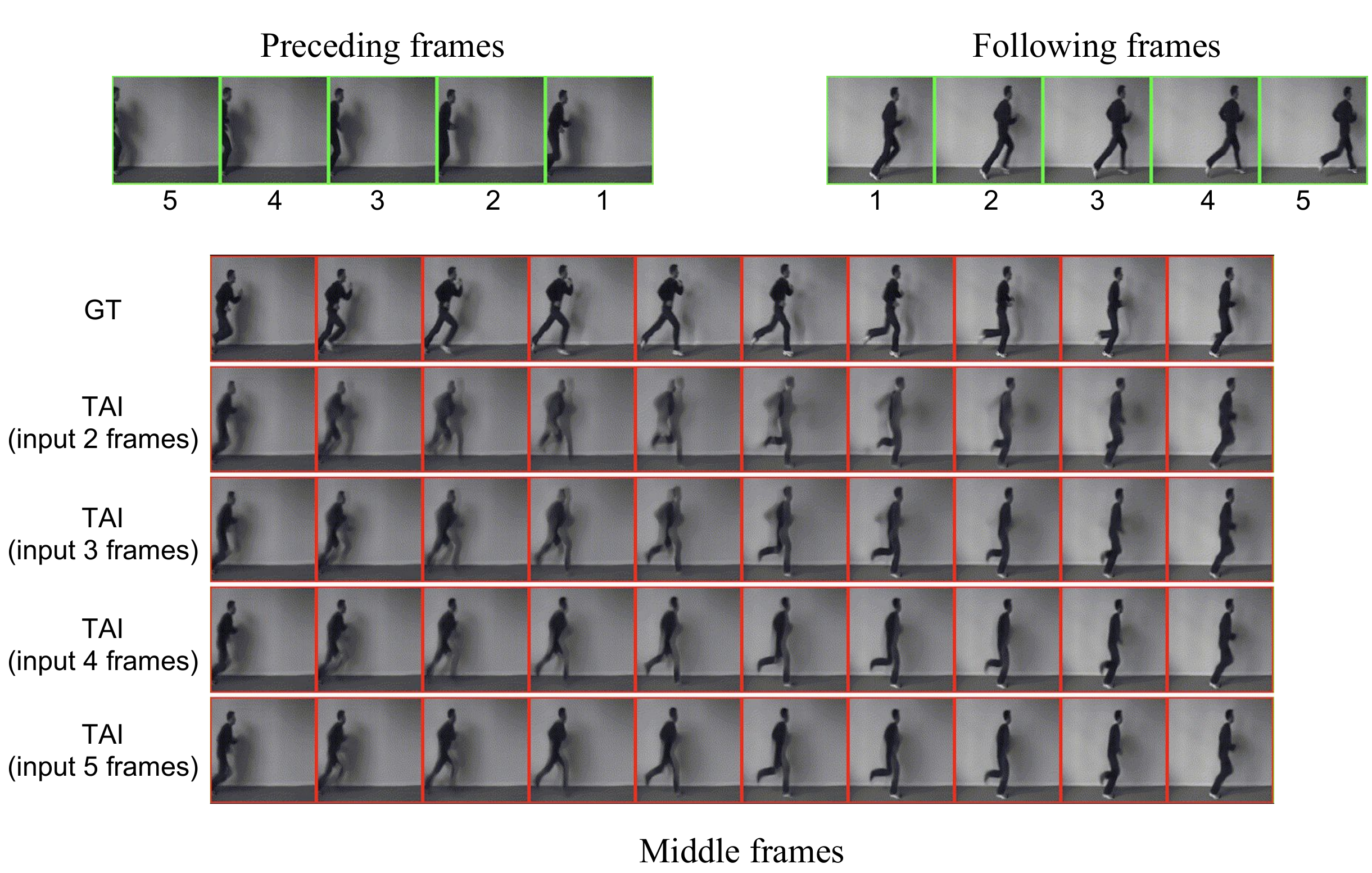}
        \caption{}
    \end{subfigure}
    \begin{subfigure}[b]{\textwidth}
        \includegraphics[width=\textwidth]{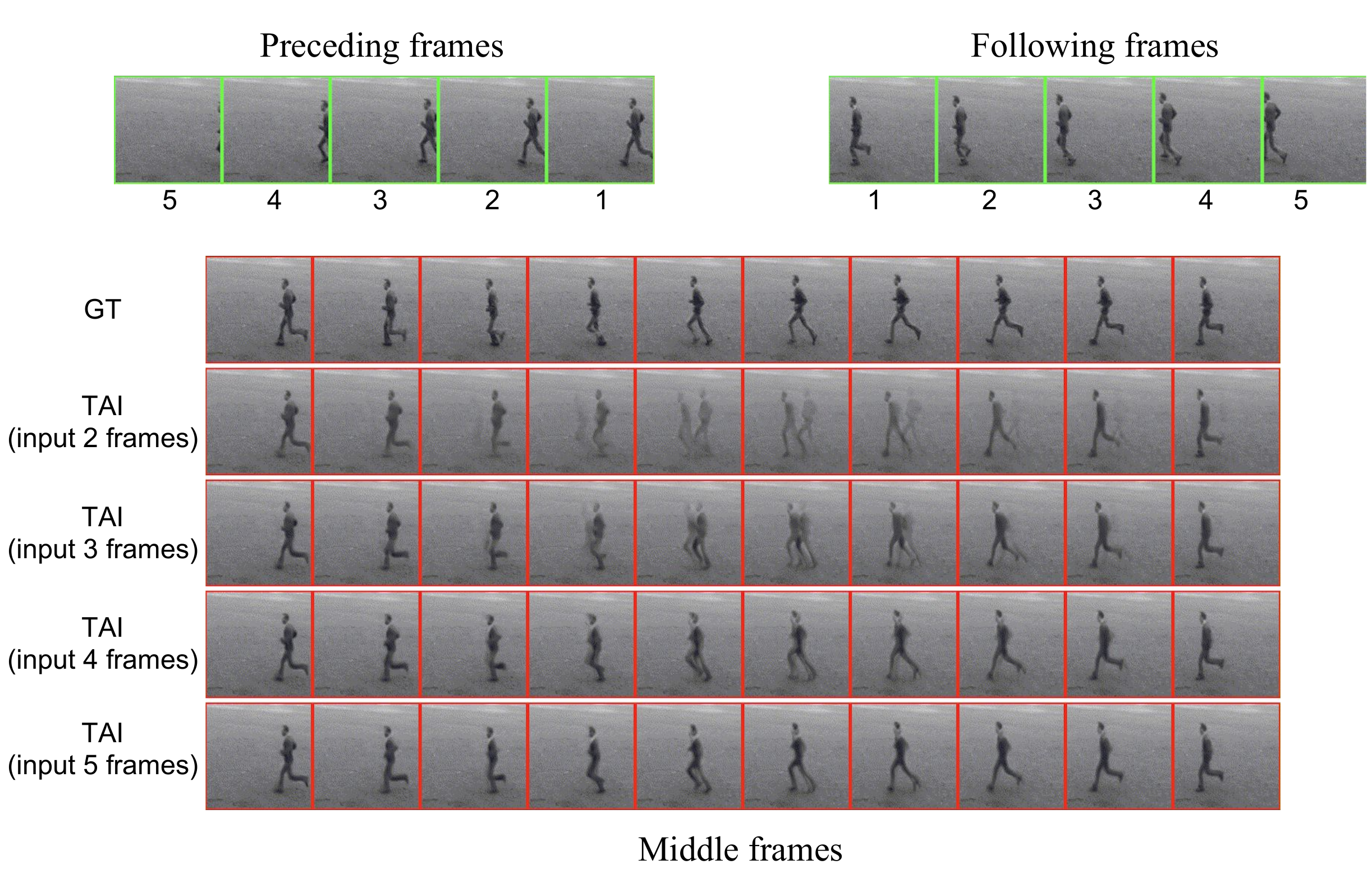}
        \caption{}
    \end{subfigure}
    \caption{Qualitative results of our trained TAI model taking in two, three, four and five preceding and following frames at test time on the KTH Actions Dataset}
    \label{fig:main_context}
\end{figure}

Comparisons in Fig.~\ref{fig:main_context} show how the final prediction changes when our model takes in a varying number of preceding and following frames at test time. When our model takes in only two frames from the preceding and following sequences, the prediction of the middle sequence fails to give one unified location for the actor due to the limited amount of available context information; however, when it takes in five frames, the final prediction contains much fewer ghosting artifacts. These results demonstrate that our TAI model can leverage the context information from multiple preceding and following frames to enhance the quality of the final prediction.

\section{Comparison of Trained Models for Variable Number of Middle Frames}
\begin{figure}[H]
	\vspace{-5pt}
	\begin{subfigure}[b]{\textwidth}
    \centering
    \includegraphics[width=0.8\textwidth]{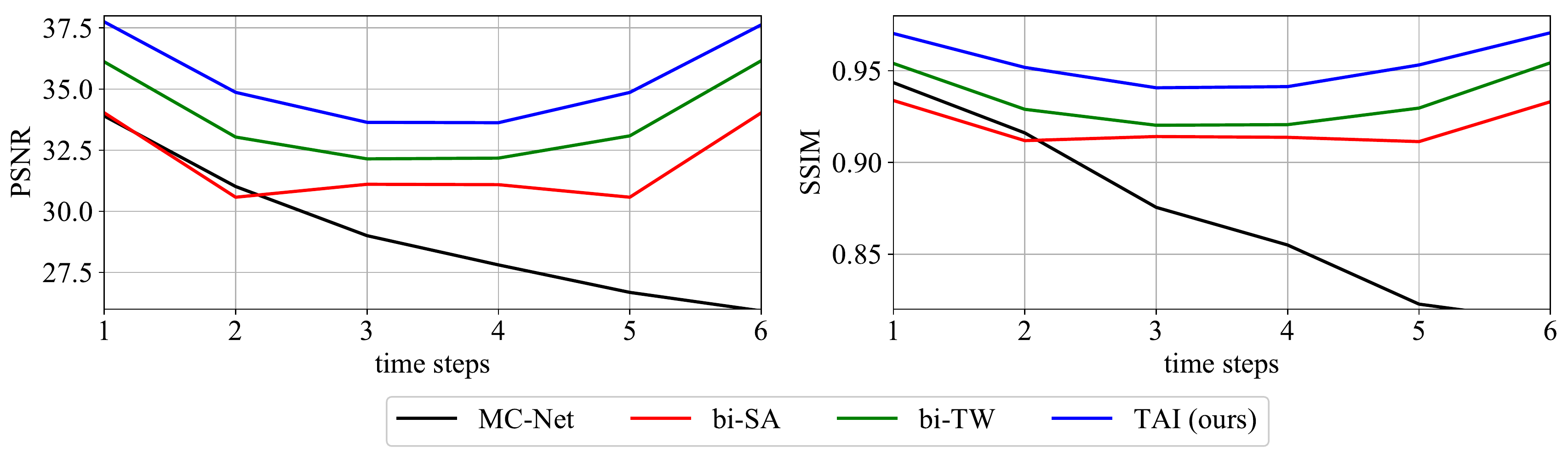}
    \end{subfigure}
    \begin{subfigure}[b]{\textwidth}
    \centering
    \includegraphics[width=0.8\textwidth]{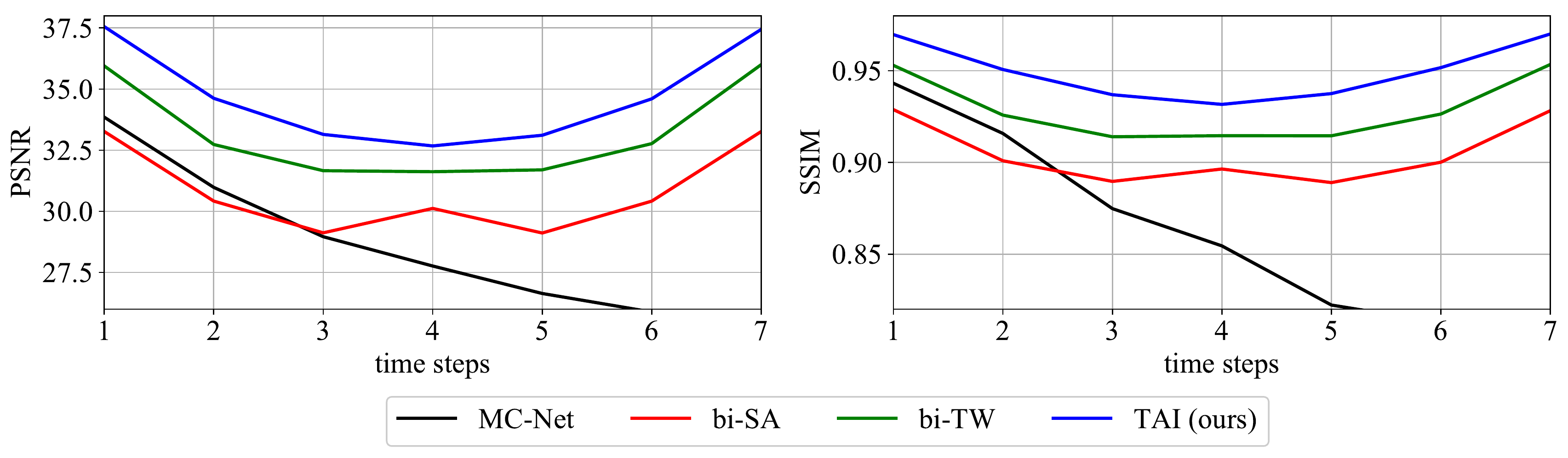}
    \end{subfigure}
    \begin{subfigure}[b]{\textwidth}
    \centering
    \includegraphics[width=0.8\textwidth]{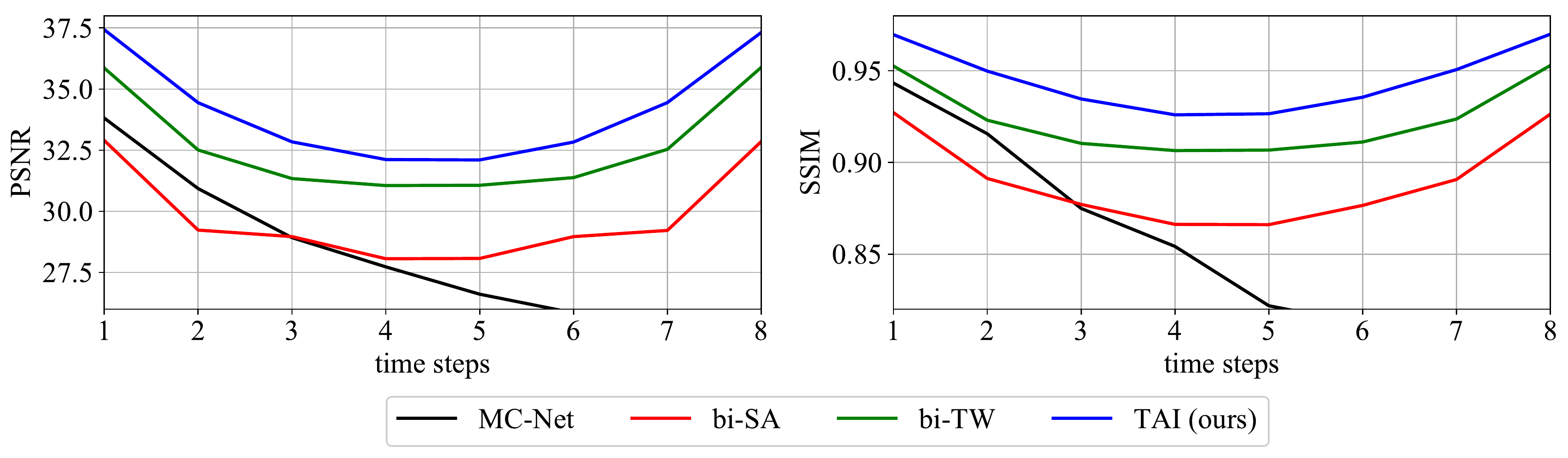}
    \end{subfigure}
    \begin{subfigure}[b]{\textwidth}
    \centering
    \includegraphics[width=0.8\textwidth]{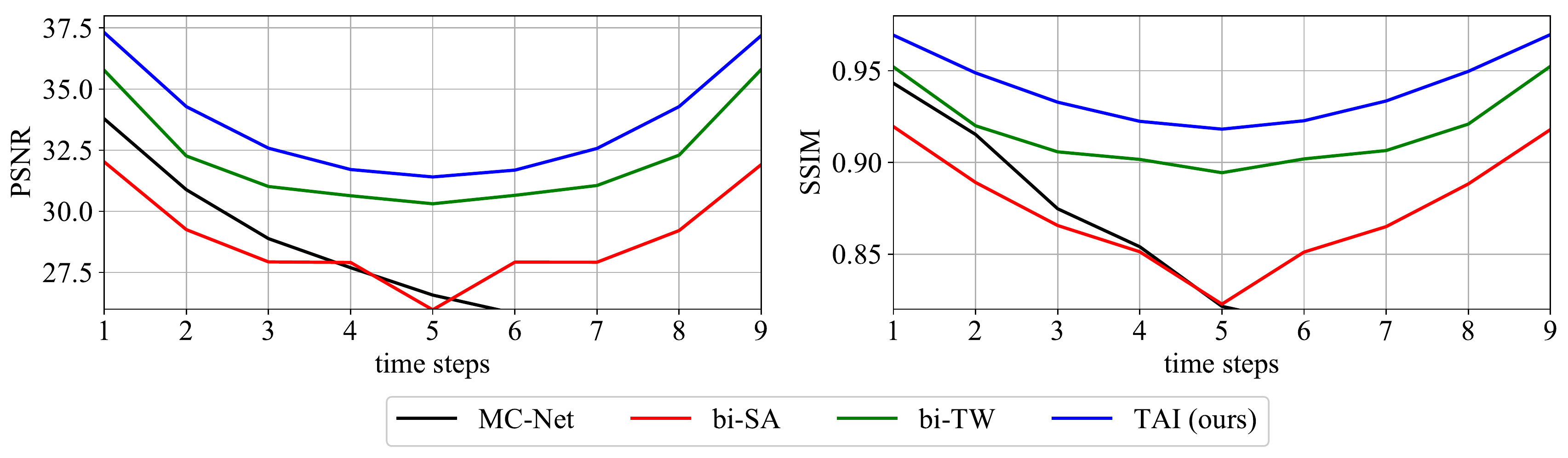}
    \end{subfigure}
    \vspace{-15pt}
    \caption{Comparison of our trained model TAI with baselines when predicting 6-9 middle frames at test time on the KTH Actions dataset. Higher PSNR and SSIM is better}
    \label{fig:diff_nums_outputs}
	\vspace{-5pt}
\end{figure}

In this section, we demonstrate that our TAI model generates higher-quality predictions of the middle frames than MC-Net, bi-SA, and bi-TW across a variable number of middle frames to inpaint. To do this, we take all four models, which were all trained to predict five middle frames, and compare their performance when predicting six, seven, eight, or nine middle frames at test time on the KTH Actions dataset. From Fig.~\ref{fig:diff_nums_outputs}, we observe that our model yields higher PSNR and SSIM values than the other models when we increase the number of outputs from six to nine at the test time. This suggests that TAI incorporates the scaled time location in a way that generalizes to a variable number of middle frames, even though it has only seen scaled time locations corresponding to five middle frames.

\section{Architecture, Training, and Evaluation Details}

\subsection{Model Architecture}
For the video prediction module, we use the same architecture as Villegas et al.~\cite{villegas2017decomposing}. As for our TAI module, the encoder is a chain of two VGG blocks~\cite{simonyan2015very} and the decoder is a chain of four VGG blocks. The kernel generation part, $\phi^{kg}_{blend}$, contains four independent sub-networks that each predict one 1D kernel. The details of each VGG block are described in Table~\ref{table:model_arch}. The inputs of decoder 2 and decoder 1 are the element-wise sum of: (i) the output of the previous decoder layer, and (ii) the residual activation from the video prediction module with matching resolution.

\begin{table}
    \begin{center}
        \caption{Architecture of TAI network}
        \label{table:model_arch}
        \scalebox{0.9}{
            \begin{tabular}{c|ccccc}
                \hline
                \hline
                block name & type& kernel size&input channels& output channels& activation\\
                \hline
                \hline
                encoder 1  & conv& 3x3& 1024& 256& RELU\\
                 &  conv& 3x3& 256& 256& RELU \\
                 &  conv& 3x3& 256& 256& RELU \\
                  \cline{2-6}
                 &  \multicolumn{5}{c}{max pooling } \\
                \hline
                encoder 2  & conv& 3x3& 256& 512& RELU\\
                 &  conv& 3x3& 512& 512& RELU \\
                 &  conv& 3x3& 512& 512& RELU \\
                   \cline{2-6}
                 &  \multicolumn{5}{c}{max pooling } \\
                \hline
                \hline
                decoder 4  & conv& 3x3& 512& 512& RELU\\
                 &  conv& 3x3& 512& 512& RELU \\
                 &  conv& 3x3& 512& 512& RELU \\
                   \cline{2-6}
                 &  \multicolumn{5}{c}{bilinear unsampling} \\
                \hline
                decoder 3  & conv& 3x3& 512& 256& RELU\\
                 &  conv& 3x3& 256& 256& RELU \\
                 &  conv& 3x3& 256& 256& RELU \\
                   \cline{2-6}
                 &  \multicolumn{5}{c}{bilinear unsampling} \\
                \hline
                decoder 2  & conv& 3x3& 256& 128& RELU\\
                 &  conv& 3x3& 128& 128& RELU \\
                 &  conv& 3x3& 128& 128& RELU \\
                   \cline{2-6}
                 &  \multicolumn{5}{c}{bilinear unsampling} \\
                \hline
                decoder 1  & conv& 3x3& 128& 64& RELU\\
                 &  conv& 3x3& 64& 64& RELU \\
                 &  conv& 3x3& 64 & 64& RELU \\
                   \cline{2-6}
                 &  \multicolumn{5}{c}{bilinear unsampling} \\
                \hline
                \hline
                kernel generation x4  & conv& 3x3& 65& 64& RELU\\
                 &  conv& 3x3& 64& 64& RELU \\
                 &  conv& 3x3& 64 & 51& RELU \\
                   \cline{2-6}
                 &  \multicolumn{5}{c}{bilinear unsampling} \\
                \hline
            \end{tabular}
        }
        \vspace{-15pt}
    \end{center}
\end{table}

\subsection{Training, Validation, and Test Set Construction}

During training, we construct minibatches by first selecting clips with $T$ frames randomly from the videos in the training set, where $T=p+m+f$, and then splitting each clip into $p$ preceding, $m$ middle, and $f$ following frames. For KTH Actions, $p=m=f=5$; for HMDB-51 and UCF-101, $p=f=4$ and $m=3$. To augment the training data, each video clip is randomly flipped horizontally or time-reversed. To validate a model, we take the first $T$ frames from each video in the validation set and split the clips into $p$ preceding, $m$ middle, and $f$ following frames.

We construct the test set differently for each dataset. For KTH Actions, we extract all clips from a sliding window of size $T'=p+m'+f$ and stride $s$ across all test videos, and then split each clip into $p$ preceding, $m'$ middle, and $f$ following frames. In our experiments, $p=f=5$, $m'=10$, and $s$ depends on the action class ($s=3$ for the running and jogging classes, and $s=m'$ for the walking, boxing, handclapping, and handwaving classes, following the stride selection process used by Villegas et al.~\cite{villegas2017decomposing}).
For HMDB-51 and UCF-101, we only evaluate each model on the first $T'$ frames of each video in the test set, where $T'=p+m'+f$, $p=f=4$, and $m'=5$. We do not evaluate on all possible clips due to the large number of test videos in these datasets.

\subsection{Training Hyperparameters}

We train the bi-SA, bi-TW, TWI and TAI models for 100,000 iterations with batch size 4. We train MC-Net for 200,000 iterations with the same batch size because it effectively receives less data per minibatch than the other models (it does not explicitly receive the following frames, unlike the other models). We use the Adam optimizer~\cite{kingma2015adam} with initial learning rate $\alpha=10^{-4}$, first decay rate $\beta_1=0.5$, and second decay rate $\beta_2=0.999$.
In the generator loss, we set the weight of the reconstruction losses $\alpha$ to 1, and the weight of the adversarial loss $\beta$ to 0.002.
We assume our discriminator can be represented as a 3-Lipschitz continuous function; thus, we apply spectral normalization~\cite{miyato2018spectral} with a Lipschitz constant of 3.
We use Xavier initialization~\cite{glorot2010understanding} for each convolutional layer and uniform initialization for each linear layer (with mean 0 and variance 0.0001 for the weights). The bias of each layer is initialized with constant 0s.


\newpage

\begin{thebibliography}{10}
\providecommand{\url}[1]{\texttt{#1}}
\providecommand{\urlprefix}{URL }
\providecommand{\doi}[1]{https://doi.org/#1}

\bibitem{borzi2003optimal}
Borzi, A., Ito, K., Kunisch, K.: {Optimal Control Formulation For Determining
  Optical Flow}. SIAM Journal On Scientific Computing  \textbf{24}(3),
  818--847 (2003)

\bibitem{chen2011image}
Chen, K., Lorenz, D.A.: {Image Sequence Interpolation Using Optimal Control}.
  Journal of Mathematical Imaging and Vision  \textbf{41}(3),  222--238 (2011)

\bibitem{cheung2008video}
Cheung, V., Frey, B.J., Jojic, N.: {Video Epitomes}. International Journal of
  Computer Vision  \textbf{76}(2),  141--152 (2008)

\bibitem{ebdelli2015video}
Ebdelli, M., Le~Meur, O., Guillemot, C.: {Video Inpainting With Short-term
  Windows: Application To Object Removal And Error Concealment}. IEEE
  Transactions on Image Processing  \textbf{24}(10),  3034--3047 (2015)

\bibitem{glorot2010understanding}
Glorot, X., Bengio, Y.: Understanding the difficulty of training deep
  feedforward neural networks. In: Proceedings of the Thirteenth International
  Conference on Artificial Intelligence and Statistics. pp. 249--256 (2010)

\bibitem{granados2012background}
Granados, M., Kim, K.I., Tompkin, J., Kautz, J., Theobalt, C.: {Background
  Inpainting For Videos With Dynamic Objects And A Free-Moving Camera}. In:
  European Conference on Computer Vision. pp. 682--695 (2012)

\bibitem{ioffe15batch}
Ioffe, S., Szegedy, C.: {Batch Normalization: Accelerating Deep Network
  Training by Reducing Internal Covariate Shift}. In: International Conference
  On Machine Learning. pp. 448--456 (2015)

\bibitem{jia2004video}
Jia, J., Tai-Pang, W., Tai, Y.W., Tang, C.K.: {Video Repairing: Inference Of
  Foreground And Background Under Severe Occlusion}. In: IEEE Conference on
  Computer Vision and Pattern Recognition (2004)

\bibitem{jia2005video}
Jia, Y.T., Hu, S.M., Martin, R.R.: {Video Completion Using Tracking And
  Fragment Merging}. The Visual Computer  \textbf{21}(8-10),  601--610 (2005)

\bibitem{kalchbrenner2016video}
Kalchbrenner, N., Oord, A.v.d., Simonyan, K., Danihelka, I., Vinyals, O.,
  Graves, A., Kavukcuoglu, K.: {Video Pixel Networks}. In: International
  Conference On Machine Learning (2017)

\bibitem{kingma2015adam}
Kingma, D.P., Ba, J.L.: {ADAM: A Method For Stochastic Optimization}. In:
  International Conference on Learning Representations (2015)

\bibitem{kuehne11hmdb}
Kuehne, H., Jhuang, H., Garrote, E., Poggio, T., Serre, T.: {HMDB: A Large
  Video Database For Human Motion Recognition}. In: IEEE International
  Conference on Computer Vision. pp. 2556--2563 (2011)

\bibitem{liu2017video}
Liu, Z., Yeh, R., Tang, X., Liu, Y., Agarwala, A.: {Video Frame Synthesis Using
  Deep Voxel Flow}. In: International Conference on Computer Vision (ICCV).
  vol.~2 (2017)

\bibitem{long2016learning}
Long, G., Kneip, L., Alvarez, J.M., Li, H., Zhang, X., Yu, Q.: {Learning Image
  Matching By Simply Watching Video}. In: European Conference on Computer
  Vision. pp. 434--450 (2016)

\bibitem{lotter2016deep}
Lotter, W., Kreiman, G., Cox, D.: {Deep Predictive Coding Networks for Video
  Prediction and Unsupervised Learning}. International Conference on Learning
  Representations  (2017)

\bibitem{mathieu2015deep}
Mathieu, M., Couprie, C., LeCun, Y.: {Deep Multi-Scale Video Prediction Beyond
  Mean Square Error}. International Conference on Learning Representations
  (2016)

\bibitem{miyato2018spectral}
Miyato, T., Kataoka, T., Koyama, M., Yoshida, Y.: {Spectral Normalization for
  Generative Adversarial Networks}. In: International Conference on Learning
  Representations (2018)

\bibitem{newson2014video}
Newson, A., Almansa, A., Fradet, M., Gousseau, Y., P{\'e}rez, P.: {Video
  Inpainting Of Complex Scenes}. SIAM Journal on Imaging Sciences
  \textbf{7}(4),  1993--2019 (2014)

\bibitem{niklaus2017bvideo}
Niklaus, S., Mai, L., Liu, F.: {Video Frame Interpolation via Adaptive
  Separable Convolution}. In: IEEE Conference on Computer Vision and Pattern
  Recognition. pp. 261--270 (2017)

\bibitem{patwardhan2007video}
Patwardhan, K.A., Sapiro, G., Bertalm{\'\i}o, M.: {Video Inpainting Under
  Constrained Camera Motion}. IEEE Transactions on Image Processing
  \textbf{16}(2),  545--553 (2007)

\bibitem{ranzato2014video}
Ranzato, M., Szlam, A., Bruna, J., Mathieu, M., Collobert, R., Chopra, S.:
  {Video (Language) Modeling: A Baseline For Generative Models Of Natural
  Videos}. arXiv preprint arXiv:1412.6604  (2014)

\bibitem{schuldt2004recognizing}
Schuldt, C., Laptev, I., Caputo, B.: {Recognizing Human Actions: A Local Svm
  Approach}. In: International Conference on Pattern Recognition. vol.~3, pp.
  32--36 (2004)

\bibitem{shen2006video}
Shen, Y., Lu, F., Cao, X., Foroosh, H.: {Video Completion For Perspective
  Camera Under Constrained Motion}. In: International Conference on Pattern
  Recognition. vol.~3, pp. 63--66 (2006)

\bibitem{simonyan2015very}
Simonyan, K., Zisserman, A.: {Very Deep Convolutional Networks for Large-Scale
  Image Recognition}. International Conference on Learning Representations
  (2015)

\bibitem{soomro2012ucf101}
Soomro, K., Zamir, A.R., Shah, M.: {UCF101: A Dataset Of 101 Human Actions
  Classes From Videos In The Wild}. CRCV-TR-12-01  (2012)

\bibitem{srivastava2015unsupervised}
Srivastava, N., Mansimov, E., Salakhudinov, R.: {Unsupervised Learning Of Video
  Representations Using LSTMs}. In: International Conference On Machine
  Learning. pp. 843--852 (2015)

\bibitem{villegas2017decomposing}
Villegas, R., Yang, J., Hong, S., Lin, X., Lee, H.: {Decomposing Motion And
  Content For Natural Video Sequence Prediction}. International Conference on
  Learning Representations  (2017)

\bibitem{wang2004image}
Wang, Z., Bovik, A.C., Sheikh, H.R., Simoncelli, E.P.: {Image Quality
  Assessment: From Error Visibility To Structural Similarity}. IEEE
  Transactions on Image Processing  \textbf{13}(4),  600--612 (2004)

\bibitem{werlberger2011optical}
Werlberger, M., Pock, T., Unger, M., Bischof, H.: {Optical flow guided TV-L1
  video interpolation and restoration}. In: International Workshop on Energy
  Minimization Methods in Computer Vision and Pattern Recognition. pp. 273--286
  (2011)

\bibitem{wexler2004space}
Wexler, Y., Shechtman, E., Irani, M.: {Space-Time Video Completion}. In: IEEE
  Conference on Computer Vision and Pattern Recognition (2004)

\end{thebibliography}

\end{document}